\documentclass[letterpaper]{article} 
\usepackage{aaai23}  
\usepackage{times}  
\usepackage{helvet}  
\usepackage{courier}  
\usepackage[hyphens]{url}  
\usepackage{graphicx} 
\usepackage{natbib}  
\usepackage{caption} 
\usepackage{algorithm}
\usepackage{algorithmic}

\usepackage{newfloat}
\usepackage{listings}
\DeclareCaptionStyle{ruled}{labelfont=normalfont,labelsep=colon,strut=off} 
\floatstyle{ruled}
\newfloat{listing}{tb}{lst}{}
\floatname{listing}{Listing}
\usepackage{xspace, bm, bbm, enumitem, color, caption, subcaption}
\usepackage{amsmath, amssymb, mathtools, amsthm}

\theoremstyle{plain}
\newtheorem{theorem}{Theorem}[section]
\newtheorem{proposition}[theorem]{Proposition}
\newtheorem{lemma}[theorem]{Lemma}
\newtheorem{corollary}[theorem]{Corollary}

\theoremstyle{remark}

\long\def\comment#1{} 

\newcommand{\xmath}[1] {\ensuremath{#1}\xspace}
\newcommand{\blmath}[1] {\xmath{\bm{#1}}}

\newcommand{\norm}[1] {\xmath{\left\| #1 \right\|}}

\newcommand{\Ab}{{\blmath A}}

\newcommand{\Ib}{{\blmath I}}

\newcommand{\pb}{{\blmath p}}
\newcommand{\qb}{{\blmath q}}
\newcommand{\rb}{{\blmath r}}

\newcommand{\ub}{{\blmath u}}
\newcommand{\vb}{{\blmath v}}
\newcommand{\wb}{{\blmath w}}
\newcommand{\xb}{{\blmath x}}
\newcommand{\yb}{{\blmath y}}
\newcommand{\zb}{{\blmath z}}

\newcommand{\Nc}{\mathcal{N}}

\newcommand{\Tc}{\mathcal{T}}

\newcommand{\Rd}{{\mathbb R}}

\newcommand{\Dc}{{{\mathcal D}}}

\newcommand{\Ed}{{{\mathbb E}}}

\newcommand{\zerob}{\mathbf{0}}

\newcommand{\beq}{\begin{equation}}
\newcommand{\eeq}{\end{equation}}
\newcommand{\beqa}{\begin{eqnarray}}
\newcommand{\eeqa}{\end{eqnarray}}

\newcommand{\indicator}[1]{\mathbbm{1}_{\{#1\}}}

\graphicspath{{figs/}}

\title{Magnitude and Angle Dynamics in Training Single ReLU Neurons}
\author{
    Sangmin Lee\equalcontrib\textsuperscript{\rm ,1},
    Byeongsu Sim\equalcontrib\textsuperscript{\rm ,1}, 
    Jong Chul Ye\textsuperscript{\rm 2, 1}
}
\affiliations{
    \textsuperscript{\rm 1} Dept. of Mathematical Sciences \\
    \textsuperscript{\rm 2} Kim Jaechul Graduate School of AI \\
    KAIST, Daejeon, South Korea\\
    
    \{ \tt{leeleesang, byeongsu.s, jong.ye} \}@kaist.ac.kr
}

\usepackage{bibentry}

\begin{document}
\allowdisplaybreaks
\maketitle

\begin{abstract}
    To understand learning the dynamics of deep ReLU networks, we investigate the dynamic system of gradient flow $\wb(t)$ by decomposing it to magnitude $\norm{\wb(t)}$ and angle $\varphi(t) := \pi - \theta(t)$ components.
    In particular, for multi-layer single ReLU neurons with spherically symmetric data distribution and the square loss function, we provide upper and lower bounds for magnitude and angle components to describe the dynamics of gradient flow.
    Using the obtained bounds, we conclude that small scale initialization  induces slow convergence speed for deep single ReLU neurons.
    Finally, {by exploiting the relation of gradient flow and gradient descent, we extend our results to the gradient descent approach.}
    All theoretical results are verified by experiments.
\end{abstract}

\section{Introduction}

To understand the underlying principles of deep neural networks (DNNs), gradient flow has been investigated in various studies. Due to complicate structure of DNNs, researchers have considered simple models like linear networks \citep{hanin2019complexity, hanin2019deep, hanin2019surprisingly},
linear diagonal networks \citep{nacson2022implicit},
two-layer ReLU networks \citep{belkin2020two, vardi2021implicit, chen2022hardness}, etc.
Surprisingly, studying even single ReLU neurons reveals some interesting properties of gradient flow that provide insights for understanding DNNs \citep{du2017convolutional, kalan2019fitting, yehudai2020learning, vardi2021learning, wu2022learning}.
For example, \citet{vardi2021implicit} show that there is no explicit regularization for single ReLU neurons, which was expected to exist in DNNs \citep{gunasekar2017implicit, soudry2018implicit}.




Another perspective of understanding DNNs is studying the training dynamics of gradient flow with its magnitude and angle. 
There are some previous works that considered the  angle between the learning parameter and true one to prove global convergence in convolutional single ReLU filter or two-layer ReLU networks \citep{brutzkus2017globally, du2017convolutional, tian2017analytical}. 
\citet{brutzkus2019larger} studied clustering dynamics, providing a constant upper bound of the angle between the convergent point and clustered data.
Similarly, the growth of norm of the parameters is also widely studied to investigate the margin maximization and implicit bias of gradient flow \citep{soudry2018implicit, ji2018gradient, ji2018risk, lyu2019gradient}.
Although those previous works revealed some properties of gradient flow and convergence to the global minimum, there is still no precise description on magnitude and angle dynamics of gradient flow.
For example, it is not known whether gradient flow has oscillation, spiral shape, or some periodic behavior during training.

In this paper, we propose a method to address these questions in a simple case of deep ReLU networks: one single neurons for hidden layers. Training over spherically symmetric data distribution with the square loss function, we decompose the gradient flow to the magnitude $\norm{\wb(t)}$ and the angle $\varphi(t)$ components and provide their upper and lower bounds. 
By investigating the obtained bounds, we show the effect of depth in single ReLU neurons. 

In addition, we also deal with gradient  descent, which is the most popular optimization algorithm used in practice. Even though it is the implementation of gradient flow, there is no previous results explicitly elucidating the relation between gradient flow and gradient descent.
We provide a novel approximation of gradient descent from a specific form of gradient flow,  connecting gradient flow and gradient descent in optimization theory. 


Our main contributions are summarized below :
\begin{itemize}

    \item We decompose the gradient flow $\wb(t)$ into the magnitude $\norm{\wb(t)}$ and the angle $\varphi(t)$ components, and induce their coupled dynamic systems. Investigating the dynamic systems, we provide a simple proof for global convergence of multi-layer single ReLU neurons.
    
    \item We derive the upper and lower bounds of both magnitude and angle to describe the dynamics of the gradient flow. These bounds show that both components converge exponentially fast.
    As a side benefit, we find that small norm initialization is not recommended for deep ReLU networks due to slow convergence speed, while it accelerates for one-layer ReLU neurons. We also show that the magnitude of deep single ReLU neurons tends to have monotonic behavior.
    
    \item We discovered a relation between the gradient flow and gradient descent. More precisely, for a gradient flow with the form of $g(e^{-ct})$, we show that the gradient descent at step $T$ with step size $\eta$ is approximated by $g((1-c\eta)^T)$. Using this, we generalize our obtained bounds of magnitude and angle on the gradient flow to the associated gradient descent.
    
    \item We verify our theory by numerical experiments.
    
\end{itemize}

\section{Preliminaries} \label{sec: preliminary}
\paragraph{Notation.}
Throughout this article, lowercase, boldface lowercase, and boldface uppercase letters denote scalars, vectors, and matrices, respectively. The identity matrix is denoted by $\Ib$. $\Ab^\top$ and $\vb^\top$ denote the transpose of a matrix $\Ab$ and a vector $\vb$. Euclidean norm of a vector is denoted by $\norm{\cdot}$. For a positive integer $m$, we denote $[m] =\{1,2,\cdots,m\}$. ReLU activation function is denoted by $\sigma(x):=\max\{0, x \}$. We denote the indicator function by
\begin{align*}
    \indicator{c} := \begin{cases}
    	1, \qquad \text{if } c \text{ is True}, \\ 0, \qquad \text{if } c \text{ is False.}
    \end{cases}
\end{align*}

\paragraph{Data distribution.}

We consider training data that follows the standard Gaussian distribution $\Nc(\zerob, \Ib)$, which is a popular assumption in existing works \citep{tian2017analytical, brutzkus2017globally, du2017convolutional, arjevani2021analytic}.

We first introduce some covariance matrices of the truncated Gaussian distribution in the following lemma, which is the key ingredient of this paper to derive differential equations.
\begin{lemma} \label{lem: two indicators}
    Suppose $\xb \sim \Nc(\zerob, \Ib)$ and let $\ub, \vb$ be two distinct unit vectors in $\Rd^d$. Then,
    \begin{align*}
        \Ed_{\xb} [\indicator{\xb^\top\ub >0} \xb\xb^\top ] &= \frac12 \Ib, \\
        \Ed_{\xb} [ \indicator{\ub^\top\xb>0}\indicator{\vb^\top\xb>0}\xb\xb^\top]
        &=\frac12 \left(1 - \frac{\theta}{\pi}\right) \Ib \\
        + \frac{1}{2\pi\sin\theta} [-\cos\theta \; (\ub&\ub^\top + \vb\vb^\top) +\vb\ub^\top +\ub\vb^\top]
    \end{align*}
    where $\theta$ is the angle between $\ub$ and $\vb$.
    In particular, we have
    \begin{align*}
        \Ed_{\xb} \left[ \sigma(\ub^\top\xb) \sigma(\vb^\top\xb) \right] = \frac12 \left(1 - \frac{\theta}{\pi}\right)\cos \theta + \frac{1}{2\pi} \sin \theta.
    \end{align*}
\end{lemma}

The covariance matrix of data from a sector of Gaussian distribution is prescribed by the directions describing the sector.
Especially, the second form of $\Ed_{\xb} [\indicator{\xb^\top\ub >0} \xb\xb^\top ]$ reflects that its eigenvectors corresponding to the largest and the smallest eigenvalues are $\ub+\vb$ and $\ub-\vb$, which are perpendicular.
The Gaussian assumption on training data can be weaken to any spherically symmetric distribution. In Appendix \ref{app: uniform sphere}, we derive the similar result of Lemma \ref{lem: two indicators} which is equivalent up to constant multiple. For simplicity, we only consider the standard Gaussian distribution in the main text.

\paragraph{Training detail.}
We consider a realizable setting, where the label $y$ of each training data $\xb$ is generated from a ground-truth target network $f^*$. Our goal is then to train a network $f$ to converge $f^*$. $f$ and $f^*$ have the same structure, which is one-layer or multi-layer with one ReLU neuron on each hidden layer. Specifically, a {\em one-layer single ReLU neuron} is defined by $f_\wb(\xb) = \sigma(\wb^\top\xb)$, and an {\em $(m+1)$-layer single ReLU neuron} is defined by $f_{\wb, v_1, \cdots, v_m}(\xb) = v_1 \sigma( v_2 \sigma( \cdots \sigma(v_m\sigma(\wb^\top\xb))\cdots))$ for $m \ge 1$.
We take the mean square loss function, and consider the population loss defined by
\begin{align} \label{eq: loss function}
    L(\wb) := \frac12 \ \Ed_{\xb\sim\Nc(\zerob, \Ib)} [(f(\xb) - f^*(\xb))^2].
\end{align}
The subscript of $\Ed_{\xb\sim\Nc(\zerob,\Ib)}$ is usually omitted.
Note that \eqref{eq: loss function} has the unique global minimum  (Proposition \ref{prop: unique global minimum}).

\paragraph{Gradient method.}
We consider a gradient flow to minimize the loss \eqref{eq: loss function}.
One may be concerned about whether gradient flow is well defined, since ReLU is not differentiable at $x=0$. However, this obstacle is known to be solved by adjusting a subgradient constant to $\sigma'(0)$ in $[0,1]$ \citep{yehudai2020learning}. For simplicity, we define $\sigma'(0)=0$ and consider the derivative of ReLU function as $\indicator{x>0}$. Then a gradient flow initialized at $\wb_0$ is well-defined, and it is a unique solution of the following differential equation :
\begin{align} \label{eq: gradient flow}
    \frac{d}{dt} \wb(t) = -\frac{\partial}{\partial\wb}L(\wb(t)),
    \qquad \wb(0) = \wb_0.
\end{align}
Similarly, gradient descent with learning rate $\eta$ at step $T$ is defined by
\begin{align} \label{eq: gradient descent}
    \wb(T+1) = \wb(T) -\eta \frac{\partial}{\partial\wb}L(\wb(T)),
    \qquad \wb(0) = \wb_0.
\end{align}

\section{Dynamics of gradient flow} \label{sec: GF}
\subsection{Warm up: one-layer single ReLU neurons} \label{sec: one layer ReLU neuron}
We start with observing the gradient flow of one-layer single ReLU neurons. In this case, the loss function \eqref{eq: loss function} and gradient flow \eqref{eq: gradient flow} are given by
\begin{align}
    L(\wb) &= \frac12 \Ed[ (\sigma(\wb^\top\xb)-\sigma(\wb^{*\top}\xb) )^2], \notag \\
    \dot \wb &= - \Ed[ \indicator{\wb^\top\xb>0} (\sigma(\wb^\top\xb) - \sigma(\wb^{*\top}\xb))\xb ] \notag \\ 
    &= -\Ed [\indicator{\wb^\top\xb>0}\sigma(\wb^\top\xb) \xb] + \Ed[\indicator{\wb^\top\xb>0}\sigma(\wb^{*\top}\xb) \xb]\notag \\
    &= -\Ed[\indicator{\wb^\top\xb>0}\xb\xb^\top]\wb \notag\\
    &\qquad + \Ed[\indicator{\wb^\top\xb>0}\indicator{\wb^{*\top}\xb>0}\xb\xb^\top] \wb^* \label{eq: gradient flow single neuron}.
\end{align}

From Lemma \ref{lem: two indicators} and the assumption $\xb\sim\Nc(\zerob, \Ib)$, we can derive the dynamical system of one-layer single ReLU neurons. Let $\theta(t)$ be the angle between $\wb(t)$ and $\wb^*$, and define $\varphi(t) := \pi - \theta(t)$, the supplementary angle of $\theta(t)$. Then, we can decompose the gradient flow $\wb(t)$ into two components: the magnitude $\norm{\wb(t)}$ and the angle $\varphi(t)$.

\begin{proposition}[The dynamical system of one-layer single ReLU neurons] \label{prop: ODE single neuron} 
    Consider a gradient flow $\wb(t)$ of a one-layer single ReLU neuron given in \eqref{eq: gradient flow single neuron}.
    Let $\theta(t) := \arccos \left( \frac{\wb(t)^\top\wb^*} {\norm{\wb(t)}\norm{\wb^*}}\right)$ and define $\varphi(t) := \pi - \theta(t)$. Then, the gradient flow $\wb(t)$ initialized at $\wb_0$ is a unique solution of the following differential equation :
    \begin{align} \label{eq: gradient flow vector form one layer}
        \frac{d}{dt} \wb = -\frac12 \wb + \frac{\varphi}{2\pi}\wb^*
        + \frac{\sin \varphi}{2\pi} \frac{\norm{\wb^*}}{\norm{\wb}}\wb,
        \qquad \wb(0)=\wb_0 .
    \end{align}
    Moreover, its magnitude $\norm{\wb(t)}$ and angle $\varphi(t)$ satisfy the following system of differential equations:
    \begin{align}
        \frac{d}{dt}\norm{\wb} &= -\frac12 \norm{\wb} + \frac{1}{2} \norm{\wb^*} \frac{\sin \varphi - \varphi \cos \varphi}{\pi},  \label{eq: norm w} \\
        \frac{d\varphi}{dt} &= \frac{1}{2} \frac{\norm{\wb^*}}{\norm{\wb}} \frac{\varphi \sin \varphi}{\pi} \label{eq: angle phi}
    \end{align}
    with initial conditions $\norm{\wb(0)} = \norm{\wb_0}$ and $\varphi(0) = \varphi_0$.
\end{proposition}
All missing proofs in this section are written in Appendix \ref{app: proof of sec one layer}.
Note that analogous statement of \eqref{eq: gradient flow vector form one layer} is also given in \cite[Theorem1]{tian2017analytical}. However, we separately focus on magnitude and angle to elucidate the dynamics of gradient flow. As shown in the following theorem, we can easily conclude that $\varphi(t)$ strictly increases to $\pi$, and $\wb(t)$ converges to the global minimum $\wb^*$.

\begin{theorem}[Global convergence of one-layer single ReLU neurons] \label{thm: global convergence one layer}
    Consider the dynamical system of a one-layer single ReLU neuron given in Proposition \ref{prop: ODE single neuron} with initial conditions $0< \varphi_0<\pi$ and $\norm{\wb_0}\neq 0$. Then, angle $\varphi(t)$ strictly increases to $\pi$ and gradient flow $\wb(t)$ converges to the global minimum $\wb^*$.
\end{theorem}
\begin{proof}
    It is enough to show that $\varphi(\infty) = \pi$ and $\norm{\wb(\infty)} = \norm{\wb^*}$. 
    We first prove $\varphi(t)$ strictly increases and converges to $\pi$. Since $ \varphi \sin \varphi >0 $ on $\varphi\in(0,\pi)$, we get $\frac{d\varphi}{dt}>0$ from \eqref{eq: angle phi}.
    Then, $\varphi(t)$ strictly increases and has an upper bound $\pi$ (from definition), thus converges to some value. 
    Let $\tilde \varphi$ be the convergent value. Then it should satisfy $\frac{d\varphi}{dt}|_{\varphi=\tilde\varphi} =0$, which implies 
    $ \tilde\varphi \sin \tilde \varphi= 0$. Since $0<\tilde\varphi\le \pi$, we conclude $\tilde\varphi = \pi$.
    Now, we prove the magnitude part. We know $\varphi(\infty)=\pi$, and we also know $\frac{d\norm{\wb(t)}}{dt}=0$ at $t=\infty$. Therefore, we get $\frac{d\norm{\wb(t)}}{dt} \big|_{t=\infty} = -\frac12 \norm{\wb(\infty)} + \frac12 \norm{\wb^*} = 0$ and conclude that $\norm{\wb(\infty)} = \norm{\wb^*}$.
\end{proof}


Another notable result is the correlation between $\varphi(t)$ and $\wb(t)$ described by \eqref{eq: norm w}.
Note that Theorem \ref{thm: global convergence one layer} shows that $\varphi(t)$ strictly increases to $\pi$, which implies that $\frac{\sin \varphi - \varphi \cos \varphi}{\pi}$ strictly increases to $1$. The dynamics of $\norm{\wb(t)}$ highly depends on $\frac{\sin \varphi - \varphi \cos \varphi}{\pi}$ term, and we provide upper and lower bounds of magnitude $\norm{\wb(t)}$ using it.

\begin{theorem} \label{thm: norm of single neuron}
    Consider the dynamical system of a one-layer single ReLU neuron given in Proposition \ref{prop: ODE single neuron}. Let $\varepsilon_0 := 1- \frac{\sin \varphi_0 - \varphi_0 \cos \varphi_0}{\pi}>0$, which is a constant determined from initialization. Then, the magnitude of gradient flow is bounded by
    \begin{align*}
        (1-\varepsilon_0)(1&-e^{-\frac12 t})\norm{\wb^*} + \norm{\wb_0}e^{-\frac12 t} \\
        &< \norm{\wb(t)} <
        (1-e^{-\frac12 t})\norm{\wb^*} + \norm{\wb_0}e^{-\frac12 t}.
    \end{align*}
    For the special case that the norm of initialization point is sufficiently small, i.e. $\norm{\wb_0}\approx 0$, the bounds are given by
    \begin{align} \label{eq: norm bound single neuron}
        (1-\varepsilon_0) (1-e^{-\frac12 t}) \norm{\wb^*} 
        < \norm{\wb(t)} < 
        (1-e^{-\frac12 t}) \norm{\wb^*}.
    \end{align} 
\end{theorem}

Recall that Theorem \ref{thm: global convergence one layer} only guarantees the convergence to the  global minimum, but nothing about the convergence speed. For a complementary result of Theorem \ref{thm: global convergence one layer}, Theorem \ref{thm: norm of single neuron} provides how fast $\norm{\wb(t)}$ converges to $\norm{\wb^*}$.
Note also that these bounds become tighter for smaller $\varepsilon_0$, which follows from the small value of $\pi - \varphi(t)$. Therefore, the angle $\varphi(t)$ decides the tightness of the bounds in Theorem \ref{thm: norm of single neuron}, and we are interested in knowing how $\varphi(t)$ is close to $\pi$.
We provide upper and lower bounds of angle $\varphi(t)$ in the following theorem.

\begin{theorem} \label{thm: angle of single neuron}
    Consider the dynamical system of a one-layer single ReLU neuron given in Proposition \ref{prop: ODE single neuron}.
    Suppose there exist constants $r,R > 0$ such that $r<\norm{\wb(t)}<R$ for all $t>0$. Then, $\varphi(t)$ is bounded by 
    \begin{align*} 
        &\pi - 2 \cot \frac{\varphi_0}{2} \; e^{-\frac{\norm{\wb^*}}{2R}\frac{\varphi_0}{\pi} t}
        < \varphi(t) \\
        &\qquad < \pi - 2\cot \frac{\varphi_0}{2}\; e^{-\frac{\norm{\wb^*}}{2r}t} + \frac23 \cot^3 \frac{\varphi_0}{2}\; e^{-\frac{3\norm{\wb^*}}{2r}t}.
    \end{align*}
    For the special case that the norm of initialization point is sufficiently small, i.e. $\norm{\wb_0}\approx 0$, the bounds are given by
    \begin{align*}
        \pi - 2 \cot \frac{\varphi_0}{2} \; e^{-\frac{\varphi_0}{2\pi} t}
        < \varphi(t) < \pi .
    \end{align*}
\end{theorem}

Similar to Theorem \ref{thm: norm of single neuron}, Theorem \ref{thm: angle of single neuron} guarantees $\varphi(t)$ converges to $\pi$ exponentially fast. However, one may wonder the existence of such constants $r$ and $R$. Fortunately, thanks to Theorem \ref{thm: norm of single neuron}, we know upper and lower bounds of $\norm{\wb(t)}$. For example, when $\norm{\wb_0}\approx 0$, we can take $r\rightarrow0^+$ and $R=\norm{\wb^*}$ from \eqref{eq: norm bound single neuron}. 

Another problem is the value of $\varphi_0$, since we do not know $\wb^*$ before training. However, there is a reasonable assumption for $\varphi_0$ over random initialization with high probability. Proposition \ref{prop: random initialization angle} tells that randomly chosen two vectors are almost orthogonal in high dimensional space. Since we encounter high dimensional data (large $d$) rather than lower dimension (small $d$), we may assume $\varphi_0 = \frac{\pi}{2}$ in Theorem \ref{thm: angle of single neuron}.


\subsection{Multi-layer single ReLU neurons} \label{sec: multi layer ReLU neuron}
Now, we extend the results of the previous section to multi-layer single ReLU neurons.
Recall that an $(m+1)$-layer single ReLU neuron is defined by $f(\xb) = v_1\sigma(v_2\cdots \sigma(v_m \sigma(\wb^{\top}\xb))\cdots)$, for $m \ge 1$.
Focusing on its special structure that has only one neuron in each hidden layer, we can deduce the following fact.


\begin{proposition} \label{prop: equivalent to linear}
    Consider a multi-layer single ReLU neuron, and suppose $v_1 >0$. Then, for $2\le k \le m$, the sign of $v_k$'s are invariant on the gradient flow.
\end{proposition}

By Proposition \ref{prop: equivalent to linear}, to avoid trivial zero neurons, we assume that all $v_k$ ($1\le k \le m$) are positive at initialization.
Then, we can remove ReLU activation $\sigma$ in hidden layers by Proposition \ref{prop: equivalent to linear}, and conclude that $f(\xb)=v_1v_2\cdots v_m\sigma(\wb^\top\xb)$.

Before we dive into deep analysis of multi layer networks, we state the {\em balanced property} introduced by \cite[Theorem 2.1]{du2018algorithmic}. Balanced property tells that $v_{k+1}^2 - v_k^2$ is constant on a gradient flow, for each $k \in [m]$. We call {\em balanced initialization} if $v_{k+1}^2 - v_{k}^2 = 0$ for all $k =0,1,\cdots, m-1$, where we define $v_0 := \norm{\wb_0}$ for convenience.
Throughout this section, we assume that the target network $f_{\wb^*, v_1^*, \cdots, v_m^*}$ is also balanced, i.e., $\norm{\wb^*}=v_1^*=v_2^*=\cdots=v_m^*>0$. Then, using the balanced property and Lemma \ref{lem: two indicators}, we can derive the dynamical system of multi-layer single ReLU neurons.

\begin{proposition}[The dynamical system of multi-layer single ReLU neurons] \label{prop: ODE multi layer}
    Consider a gradient flow of an $(m+1)$-layer single ReLU neuron with balanced initialization, i.e., $v_1(0)=\cdots=v_m(0)=\norm{\wb_0}$. Let $\theta(t) := \arccos \left( \frac{\wb(t)^\top\wb^*} {\norm{\wb(t)}\norm{\wb^*}}\right)$ and define $\varphi(t) := \pi - \theta(t)$.
    Then, the gradient flow $\wb(t)$ satisfies the following differential equation:
    \begin{align*}
        \frac{d}{dt} \wb &= -\frac12 v^{2m} \wb + \frac{1}{2} v^m v^{*m} 
        \left( \frac\varphi\pi \wb^* + \frac{v^*}{v} \frac{\sin \varphi}{\pi} \; \wb \right).
    \end{align*}
    Moreover, its magnitude $\norm{\wb(t)}$ and angle $\varphi(t)$ satisfy the following system of differential equations:
    \begin{align}
        \frac{d}{dt} \norm{\wb} = \frac{dv}{dt} &=
        -\frac12 v^m \left( v^{m+1} - v^{*m+1} \frac{\sin \varphi - \varphi \cos \varphi}{\pi}\right), \label{eq: norm v} \\
        \frac{d\varphi}{dt} &= v^{m-1}v^{*m+1} \frac{\varphi \sin \varphi }{2\pi} . \label{eq: angle deep}
    \end{align}
\end{proposition}

All missing proofs in this section are written in Appendix \ref{app: proof of sec multi layer}. 
Like Theorem \ref{thm: global convergence one layer}, we can easily prove convergence to the global minimum for multi-layer single ReLU neurons.

\begin{theorem}[Global convergence of multi-layer single ReLU Neurons]
    \label{thm: global convergence of multi layer}
    Consider the dynamical system of a multi-layer single ReLU neuron given in Proposition \ref{prop: ODE multi layer} with balanced initialization.
    If $0< \varphi_0 <\pi$ and $\norm{\wb_0}\neq 0$, $\varphi(t)$ strictly increases to $\pi$ and the gradient flow $\wb(t)$ converges to the global minimum $\wb^*$.
\end{theorem}

Now, we provide an insight about magnitude dynamics. Let $\varepsilon(t) := 1 - \frac{\sin \varphi - \varphi \cos \varphi}{\pi}>0$ which strictly decreases to zero since $\varphi(t)$ strictly increases to $\pi$ by Theorem \ref{thm: global convergence of multi layer}. Investigating \eqref{eq: norm v}, we get

\begin{align} \label{eq: norm monotonic}
    \frac{dv}{dt} < 0 \qquad \Longleftrightarrow \qquad v > v^* (1-\varepsilon)^{\frac{1}{m+1}}.
\end{align}

This implies that if $v_0$ is sufficiently large, then 
the magnitude $v(t)$ strictly decreases until $v<v^* (1-\varepsilon)^{\frac{1}{m+1}}<v^*$.
Similarly, if $v_0$ is small enough, $v(t)$ strictly increases provided $v<v^*(1-\varepsilon)^{\frac{1}{m+1}}$. Since large $m$ mitigates the effect of $\varepsilon(t)$ term in \eqref{eq: norm monotonic}, we conclude that $\frac{dv}{dt} < 0 \Longleftrightarrow v> v^*$ for sufficiently large $m$, which implies that $v(t)$ is monotonic regardless of the initialization. 
In other words,  {\em the magnitude in deep networks tends to have monotone behavior}.
We provide an example showing this property for general deep ReLU networks, in Appendix \ref{app: general deep} Figure \ref{fig: general deep}.

For specific assertions on the magnitude behavior, we derive upper and lower bounds of the magnitude for multi-layer single ReLU neurons.
However, due to the complicate form of \eqref{eq: norm v}, the bounds of magnitude $\norm{\wb(t)}$ are presented by implicit form using \emph{Gaussian hypergeometric functions} $_2F_1(a,b;c;z)$ in the following theorem.
Fortunately, for a special case $m=1$ (i.e. two-layer single ReLU neurons), we provide explicit bounds.

\begin{theorem} \label{thm: norm of multi layer}
    Consider the dynamical system of an $(m+1)$-layer single ReLU neuron in Proposition \ref{prop: ODE multi layer} with balanced initialization. Define $\varepsilon_0:= 1- \frac{\sin \varphi_0 - \varphi_0 \cos \varphi_0}{\pi} >0$ from the initial condition.
    Let $u(t ; \varepsilon)$ be the implicit function satisfies
    \begin{align*}
        &\frac{u(t; \varepsilon)^{1-m}}{(m-1)v^{*m+1}(1-\varepsilon)} \ _2F_1\left(1, \frac{1-m}{m+1}; \frac{2}{m+1}; \frac{u(t;\varepsilon)^{m+1}}{(1-\varepsilon)v^{*m+1}}\right) \\
        &-\frac{v_0^{1-m}}{(m-1)v^{*m+1}(1-\varepsilon_0)} \ _2F_1\left(1, \frac{1-m}{m+1}; \frac{2}{m+1}; \frac{v_0^{m+1}}{(1-\varepsilon_0)v^{*m+1}}\right) \\
        &=-\frac12 t
    \end{align*}
    where $_2F_1(a,b;c;z)$ denotes the Gaussian hypergeometric function. Then, $\norm{\wb(t)}=v(t)$ is bounded by
    \begin{align*}
        u(t; \varepsilon_0) < v(t) < u(t ; 0).
    \end{align*}
    For the special case $m=1$ (i.e., two-layer), the function $u(t;\varepsilon)$ is explicitly given and the magnitude is bounded by
    \begin{align*}
        v^* &\sqrt{\frac{1-\varepsilon_0}{1 - \left(1 - (1-\varepsilon_0)(\frac{v^*}{v_0})^2\right) e^{-(1-\varepsilon_0)v^{*2}t}}} \\
        &\qquad < v(t) < 
        v^* \sqrt{\frac{1}{1 - \left(1-(\frac{v^*}{v_0})^2\right) e^{-v^{*2}t}}} .
    \end{align*}
\end{theorem}

Like Theorem \ref{thm: norm of single neuron}, 
smaller $\varepsilon_0$ provides tighter bounds because $\lim_{\varepsilon\rightarrow0^+} u(t;\varepsilon) = u(t;0)$. 
Since small value of $\varepsilon_0$ is obtained from small value of $\pi - \varphi(t)$, we need to know upper and lower bounds of $\varphi(t)$ for multi-layer single ReLU neurons.

\begin{theorem} \label{thm: angle of multi layer}
    Consider the dynamical system of an $(m+1)$-layer single ReLU neuron in Proposition \ref{prop: ODE multi layer} with balanced initialization. Suppose there exist constants $r,R >0$ such that $r<\norm{\wb(t)}<R$ for all $t>0$. Then, the angle $\varphi(t)$ is bounded by
    \begin{align*}
        &\pi - 2\cot \frac{\varphi_0}{2} e^{-\frac{\varphi_0}{2\pi}r^{m-1}v^{*m+1}t }
        < \varphi(t) \\
        &< \pi - 2\cot \frac{\varphi_0}{2} e^{-\frac{1}{2}R^{m-1}v^{*m+1}t } + \frac23 \cot^3 \frac{\varphi_0}{2} e^{-\frac{3}{2}R^{m-1}v^{*m+1}t }
    \end{align*}
\end{theorem}

We have provided upper and lower bounds of magnitude $\norm{\wb(t)}$ and angle $\varphi(t)$ of gradient flow in multi-layer single ReLU neurons  through Theorem \ref{thm: norm of multi layer} and Theorem \ref{thm: angle of multi layer}, which show exponential convergence for both components. 
Furthermore, these bounds describe the dynamics of gradient flow, in the sense that we can estimate the range of gradient flow. 
While it is intractable to track the exact trajectory of the gradient flow in high dimensional parameter space, our proposed theorems assure that the gradient flow cannot have large scale oscillation or spiral move.


Moreover, our results give an insight about the initialization scale.
\eqref{eq: norm v} and \eqref{eq: angle deep} induce $\frac{d\varphi}{dt} \propto \norm{\wb}^{m-1}$ and $\frac{d\norm{\wb}}{dt} \propto \norm{\wb}^{m}$, which reveal that small values of $r, R>0$ help one-layer neurons ($m=0$) converge fast (Theorem \ref{thm: angle of single neuron}) while multi-layer neurons ($m>1$) are benefited by large $r, R>0$ (Theorem \ref{thm: angle of multi layer}).
Therefore, with balanced initialization condition, we conclude that {\em small norm initialization ($\norm{\wb_0}\approx 0$) slows down convergence speed for multi-layer case}, nevertheless it is encouraged for one-layer single ReLU neurons to fast convergence. This claim is understood as an extension of  \citet{shamir2019exponential} which only considered one-dimensional inputs.

Lastly, we leave a remark on the kernel regime (implicit bias) with initialization scale. It is known that small norm initialization reaches to rich regime \citep{maennel2018gradient, boursier2022gradient, ma2022global} while large norm initialization converges in kernel regime \citep{chizat2019lazy, moroshko2020implicit}. Combined with our result, since we do not recommend small norm initialization due to slow convergence, we propose a trade-off in initialization scale between convergence speed and implicit bias.

\section{Generalization to gradient descent} \label{sec: GD}

In this section, we propose a relation between gradient flow and gradient descent.
Specifically, if gradient flow is a function of $e^{-ct}$, we show that gradient descent with learning rate $\eta$ at step $T$ is approximated by interchanging $e^{-ct}$ term to $(1-c\eta)^T$. Its proof can be found in Appendix \ref{app: proof of sec GD}

\begin{theorem}[Approximation of gradient descent from gradient flow] 
\label{thm: relation between GF and GD}
    Consider a differential equation with its gradient flow solution $w(t)$. Suppose the solution is of the form $w(t) = g(e^{-ct})$ for some one-to-one function $g\in C^2([0,1])$.
    Then, gradient descent with learning rate $\eta \ll \frac1c$ at step $T$ is approximated by
    \begin{align} \label{eq: relation between GF and GD}
        w(T) = g((1-c\eta)^T) + O(T\eta^2).
    \end{align}
    {Furthermore, if $(g^{-1})', g''$ and $g'(x) + xg''(x)$ are bounded, then for sufficiently small $\eta>0$, we can obtain an approximation independent to $T$:}
    \begin{align*}
        w(T) = g((1-c\eta)^T) + O(\eta).
    \end{align*}
    Conversely, if a gradient descent solution is given by a function of $(1-c\eta)^T$, then we can induce gradient flow solution by changing that term to $e^{-ct}$.
\end{theorem}

Note that \citet[Theorem 3.4]{cisneros2022contraction} showed similar results in the contracting system. However, by replacing $e^{-ct}$ by $(1-c\eta)^T$, we obtain better bounds without assuming a contracting system.
Accordingly, Theorem \ref{thm: relation between GF and GD} can connect gradient flow to gradient descent.
Since all explicitly obtained bounds of magnitude $\norm{\wb(t)}$ and angle $\varphi(t)$ in Section \ref{sec: GF} are functions of $e^{-ct}$, we can directly extend their obtained bounds to gradient  descent via Theorem \ref{thm: relation between GF and GD}. For example, here is the gradient descent version of Theorem \ref{thm: angle of multi layer}.

\begin{theorem}[Gradient descent version of Theorem \ref{thm: angle of multi layer}] \label{thm: angle of multi layer GD version}
    Consider the dynamical system of an $(m+1)$-layer single ReLU neuron in Proposition \ref{prop: ODE multi layer} trained by gradient  descent with learning rate $\eta$, with balanced initialization. Suppose there exist constants $r,R>0$ such that $r<\norm{\wb(T)}<R$ for all $T>0$. Then, if learning rate satisfies $\min\{ \frac{2\pi}{\varphi_0 r^{m-1}v^{*m+1}} , \frac{2}{3R^{m-1}v^{*m+1}} \} \gg \eta>0$, $\varphi(T)$ is bounded by
    \begin{align*}
        \pi - &2 \cot \frac{\varphi_0}{2} \left(1 -\frac{\varphi_0}{2\pi} r^{m-1} v^{*m+1} \eta \right)^T
        < \varphi(T) \\
        & < \pi - 2 \cot \frac{\varphi_0}{2} \left(1 -\frac12 R^{m-1} v^{*m+1} \eta \right)^T \\
        &\qquad\quad + \frac23 \cot^3 \frac{\varphi_0}{2} \left(1 -\frac32 R^{m-1} v^{*m+1} \eta \right)^T .
    \end{align*}
\end{theorem}

This shows the power of Theorem \ref{thm: relation between GF and GD}; {\em we can directly extend results on gradient flow to gradient descent}, thus it is enough to study gradient flow only.
Moreover, there are many asymptotic results for gradient flows represented by $e^{-ct}$ form in previous works \citep{vardi2021learning, wu2022learning, yehudai2020learning, gao2021global}. For instance, Theorem 5.3 in \citet{yehudai2020learning} could be understood as the conclusion of this theorem.

The converse part of Theorem \ref{thm: relation between GF and GD} states that we can associate gradient flow solution if gradient descent is a function of $(1-c\eta)^T$. This form is also frequently appeared in gradient descent analysis \citep{du2019gradient, hu2020provable, zou2020global, nguyen2020global, vardi2021learning}, and we can induce the corresponded gradient flow equation which would be easier to investigate. 
This connection on gradient  descent enables applications in practical setting, as the following corollary shows.

\begin{corollary} \label{cor: required T to estimate angle}
    Consider the dynamical system of an $(m+1)$-layer single ReLU neuron in Proposition \ref{prop: ODE multi layer} trained by gradient  descent with learning rate $\eta$, with balanced initialization. Suppose there exists a constant $r>0$ such that $\norm{\wb(T)}>r$ for all $T>0$. Then, if learning rate satisfies $\frac{2\pi}{\varphi_0 r^{m-1}v^{*m+1}} \gg \eta >0$, we get $\pi - \varepsilon < \varphi(T) < \pi$ for given $\varepsilon>0$ if 
    \begin{align*}
        T > \frac{\log \left( \frac\varepsilon2 \tan \frac{\varphi_0}{2}\right)}
        {\log \left(1-\frac{\varphi_0}{2\pi} r^{m-1} v^{*m+1}\eta\right)}  .
    \end{align*}
\end{corollary}

Assuming $v^*$ is given and $r=\norm{\wb_0}$ is sufficiently small, and taking $\varphi_0=\frac\pi2$ according to Proposition \ref{prop: random initialization angle}, Corollary \ref{cor: required T to estimate angle} estimates iterating steps(epochs) required to converge under given threshold $\varepsilon>0$ for angle $\varphi(T)$. We can find practical merits of this corollary, in the point that we can set stopping time for training before we start. 
Other theorems in Section \ref{sec: GF} are also extended to gradient descent via Theorem \ref{thm: relation between GF and GD}, and written in Appendix \ref{app: GD}.


\begin{figure*}[ht]
	\centering
	\begin{subfigure}[b]{0.32\textwidth}
        \centering
        \includegraphics[width=\textwidth]{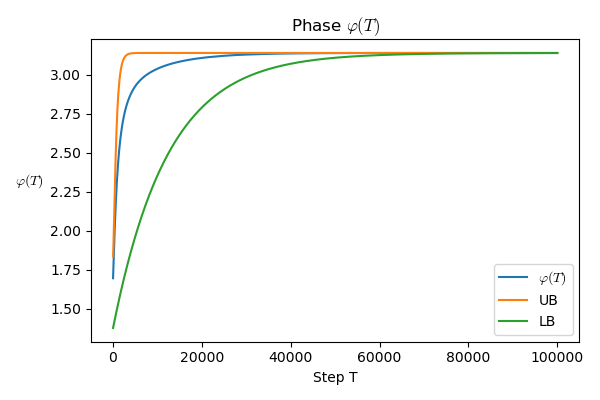}
        \caption{1L-SN}
    \end{subfigure}
    \hfill
	\begin{subfigure}[b]{0.32\textwidth}
        \centering
        \includegraphics[width=\textwidth]{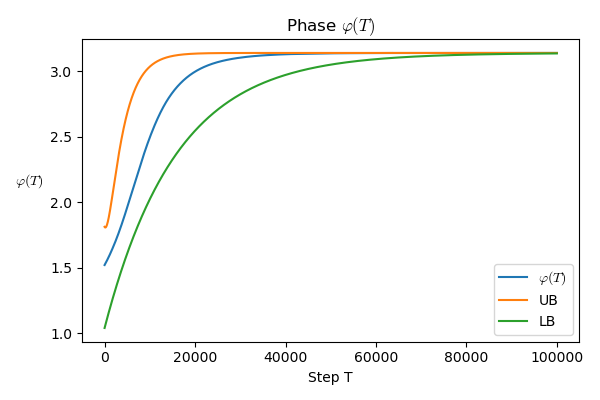}
        \caption{1L-MN}
    \end{subfigure}
    \hfill
    \begin{subfigure}[b]{0.32\textwidth}
        \centering
        \includegraphics[width=\textwidth]{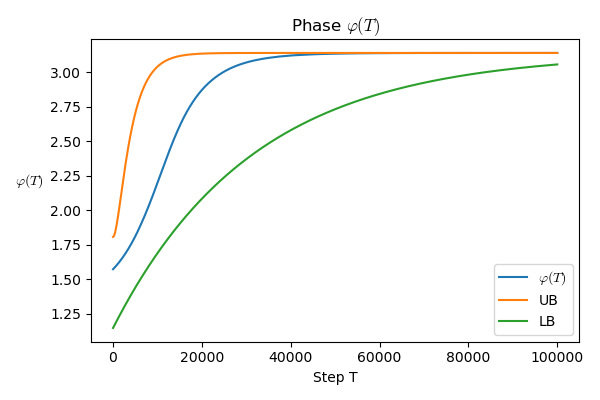}
        \caption{1L-LN}
    \end{subfigure}
    \\
    \begin{subfigure}[b]{0.32\textwidth}
        \centering
        \includegraphics[width=\textwidth]{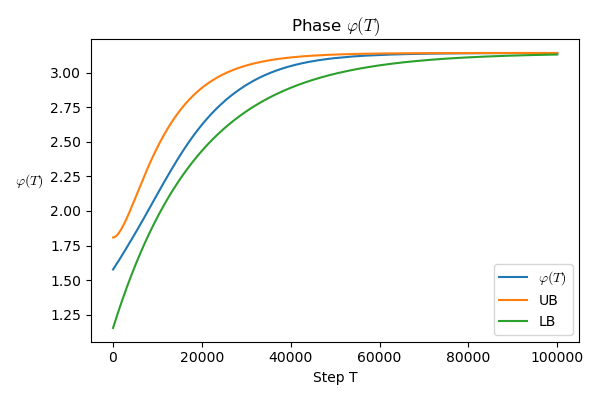}
        \caption{2L-SN}
    \end{subfigure}
    \hfill
	\begin{subfigure}[b]{0.32\textwidth}
        \centering
        \includegraphics[width=\textwidth]{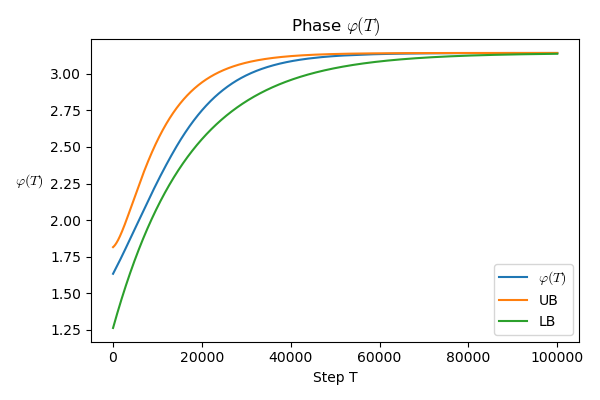}
        \caption{2L-MN}
    \end{subfigure}
    \hfill
    \begin{subfigure}[b]{0.32\textwidth}
        \centering
        \includegraphics[width=\textwidth]{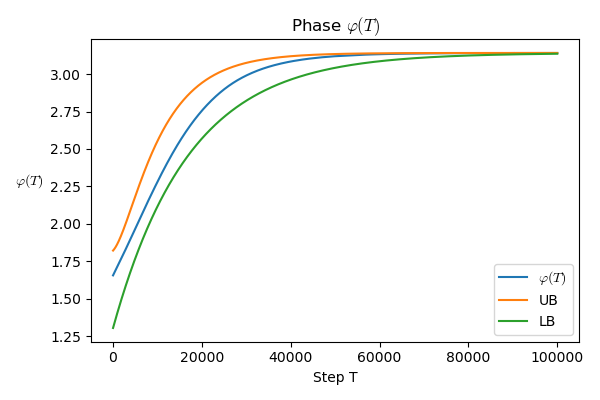}
        \caption{2L-LN}
    \end{subfigure}
    \\
    \begin{subfigure}[b]{0.32\textwidth}
        \centering
        \includegraphics[width=\textwidth]{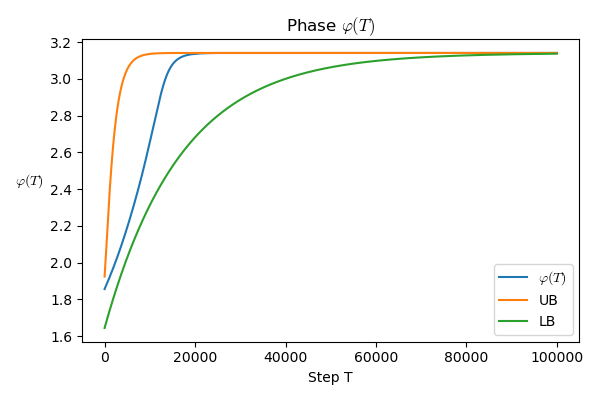}
        \caption{3L-SN}
    \end{subfigure}
    \hfill
	\begin{subfigure}[b]{0.32\textwidth}
        \centering
        \includegraphics[width=\textwidth]{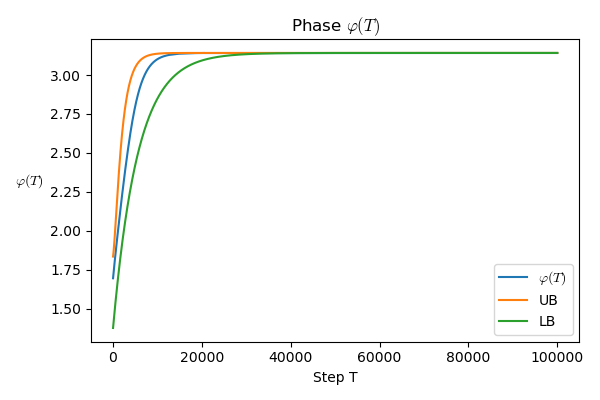}
        \caption{3L-MN}
    \end{subfigure}
    \hfill
    \begin{subfigure}[b]{0.32\textwidth}
        \centering
        \includegraphics[width=\textwidth]{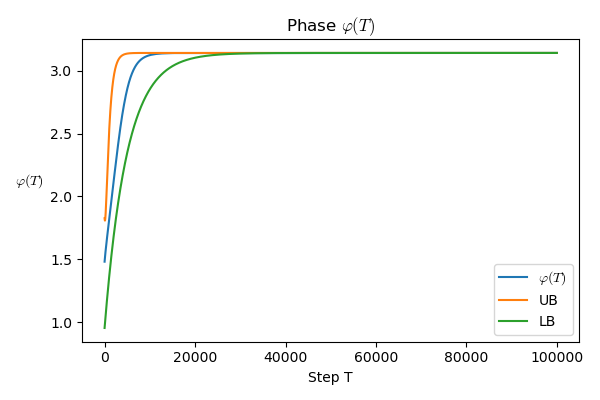}
        \caption{3L-LN}
    \end{subfigure}
    \caption{
    Angle dynamics in single ReLU neurons. First to third row represent one, two and three-layers (1L, 2L, 3L), and first to third columns represent small, middle and large norm (SN, MN, LN) initialization. We plot the angle $\varphi(T)$ (blue) with upper and lower bounds (green, orange) provided by Theorem \ref{thm: angle of single neuron GD version} and Theorem \ref{thm: angle of multi layer GD version}.
    }
    \label{fig: angle}
\end{figure*}
\section{Experiments} \label{sec: experiments}

In this section, we provide numerical results.
We construct training dataset $\{\xb_i\}_{i=1}^n \in \Rd^d$ from $\Nc(\zerob, \Ib)$ with $d=100$ and $n=10000$. The label of $\xb_i$ is generated by a target network, as described in Section~\ref{sec: preliminary}. 
The number of total iterations(epochs) is fixed to $100000$, where learning rate $\eta$ is flexible for each experiment. Detail setting is written in Appendix \ref{app: setting}.

\subsection{Angle of single ReLU neurons}

We observe angle dynamics $\varphi(T)$ for one, two and three-layer single ReLU neurons, with various initialization setting. Its upper and lower bounds are drawn according to Theorem \ref{thm: angle of single neuron GD version} and Theorem \ref{thm: angle of multi layer GD version} for one-layer and multi-layer single ReLU neurons. We consider three initialization schemes for each experiment, which are small norm($\norm{\wb_0}\approx 0$), middle norm($\norm{\wb_0}\approx \norm{\wb^*}$) and large norm($\norm{\wb_0} > 2\norm{\wb^*}$) initialization. The values of $r$ and $R$ are suitably determined for each case, which are all written in Appendix \ref{app: setting}. The training dynamics of angle $\varphi(T)$ is shown in Figure \ref{fig: angle}.

Note first that the magnitude of initialization($\norm{\wb_0}$) indeed affects the convergence speed of $\varphi(t)$. Figure \ref{fig: angle}(a) shows small norm initialization forces $\varphi(t)$ converges very fast, as \eqref{eq: angle phi} implies $\frac{d\varphi}{dt} \propto \frac{1}{\norm{\wb(t)}}$.
However, Figure \ref{fig: angle}(g) shows small norm initialization slows down the convergence speed of $\varphi(t)$, as \eqref{eq: angle deep} implies $\frac{d\varphi}{dt} \propto {\norm{\wb(t)}^{m-1}}$. 
This verifies the aforementioned claim in Section \ref{sec: GF}: {\em small-norm initialization is not preferred in terms of convergence speed for deeper layers} since $\frac{d\varphi}{dt} \propto v^{m-1}$.
More precisely, for two-layer single ReLU neurons ($m=1$), $\varphi(t)$ is independent with magnitude $\norm{\wb(t)}$ \citep{maennel2018gradient}.
This might be understood as a boundary between one-layer networks and deep networks, like small norm initialization accelerates the convergence speed of one-layer ($m=0$) while it decelerates for multi-layers ($m\ge2$).

\begin{figure*}[ht]
	\centering
	\begin{subfigure}[b]{0.32\textwidth}
        \centering
        \includegraphics[width=\textwidth]{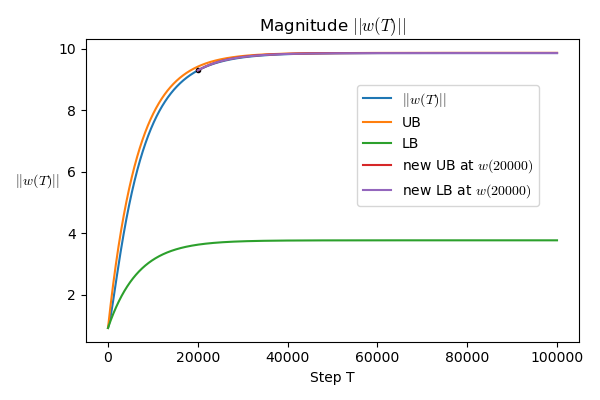}
        \caption{1L-SN}
    \end{subfigure}
    \hfill
	\begin{subfigure}[b]{0.32\textwidth}
        \centering
        \includegraphics[width=\textwidth]{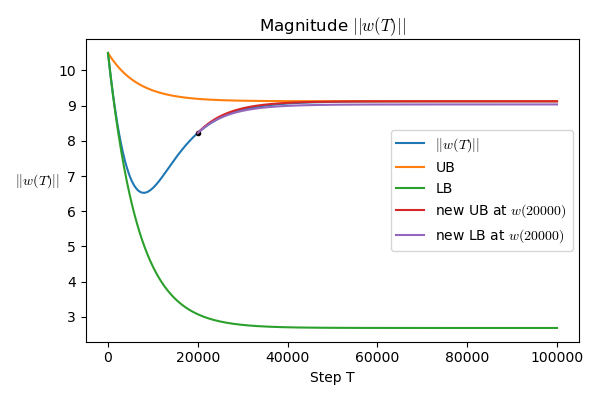}
        \caption{1L-MN}
    \end{subfigure}
    \hfill
	\begin{subfigure}[b]{0.32\textwidth}
        \centering
        \includegraphics[width=\textwidth]{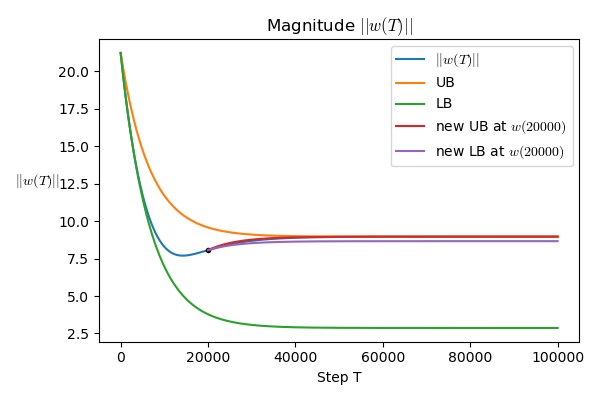}
        \caption{1L-LN}
    \end{subfigure}
    \\
    \begin{subfigure}[b]{0.32\textwidth}
        \centering
        \includegraphics[width=\textwidth]{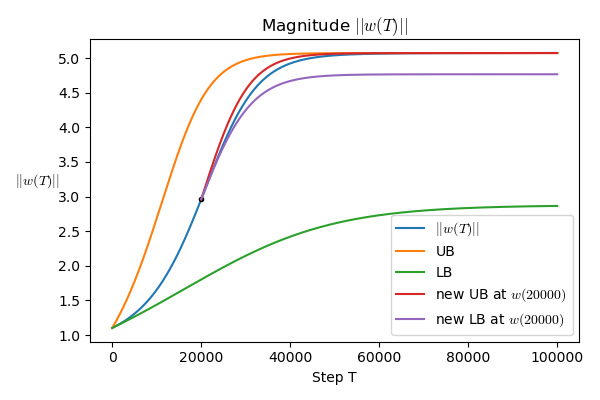}
        \caption{2L-SN}
    \end{subfigure}
    \hfill
	\begin{subfigure}[b]{0.32\textwidth}
        \centering
        \includegraphics[width=\textwidth]{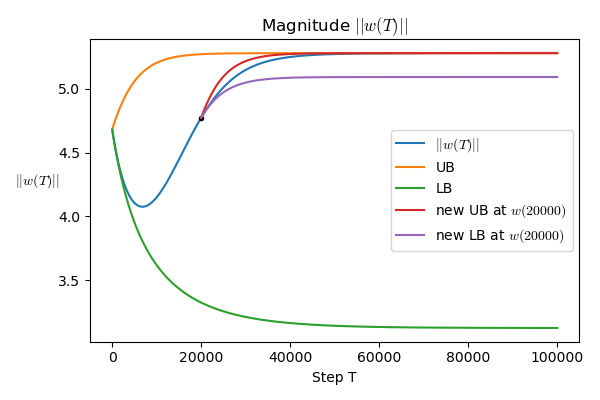}
        \caption{2L-MN}
    \end{subfigure}
    \hfill
	\begin{subfigure}[b]{0.32\textwidth}
        \centering
        \includegraphics[width=\textwidth]{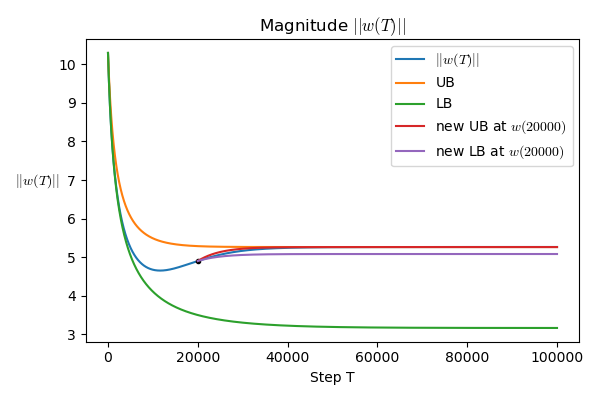}
        \caption{2L-LN}
    \end{subfigure}
    \caption{
    Magnitude dynamics in single ReLU neurons. First and second row represent one and two-layers (1L, 2L), and first to third columns represent small, middle and large norm (SN, MN, LN) initialization. 
    We plot the magnitude $\norm{\wb(t)}$ (blue) with upper and lower bounds (green, orange) provided by Theorem \ref{thm: norm of single neuron GD version} and Theorem \ref{thm: norm of two layer GD version}. We also provide new upper and lower bounds (purple, red) initialized at $\wb(20000)$ (black dot).
    }
    \label{fig: magnitude}
\end{figure*}
\subsection{Magnitude of single ReLU neurons}

Here we exhibit similar experimental results for magnitude $\norm{\wb(T)}$. Due to implicit form of bounds in 
Theorem \ref{thm: norm of multi layer} for $m\ge2$, we investigate the dynamics of magnitude for one and two-layer single ReLU neurons (i.e., $m=0, 1$). The training dynamics are shown in Figure \ref{fig: magnitude}.

One can notice a problem in Figure \ref{fig: magnitude}: lower bounds (green curves) are too loose to describe the dynamics of $\norm{\wb(T)}$. 
More precisely, the lower bound tends to follow $\norm{\wb(T)}$ in the beginning of training process, while the upper bound approximates the end of training.
This is caused by $\varepsilon_0 := 1 - \frac{\sin\varphi_0 - \varphi_0\cos\varphi_0}{\pi}$ defined in Theorem \ref{thm: norm of single neuron GD version} and Theorem \ref{thm: norm of two layer GD version}. The lower bound becomes tighter for small $\varepsilon_0$ value, which is reduced as $\varphi(T)$ converges to $\pi$. To resolve this obstacle, it is recommended to reconstruct upper and lower bounds after $\varphi(T)$ goes sufficiently close to $\pi$ (thus $\varepsilon(T)$ is small enough). For example, we set $\wb(20000)$ as an initialization of new bounds, which are denoted by black dots in Figure \ref{fig: magnitude}. Constructed new upper and lower bounds(red, purple) provide more tighter range for magnitude $\norm{\wb(T)}$. We also provide several bounds induced from distinct initialization points on Figure \ref{fig: several init} in Appendix \ref{app: several init}, which shows that latest bounds are the tightest bounds for magnitude $\norm{\wb(T)}$.

We conclude this section with discussion on the shape of magnitude curve $\norm{\wb(T)}$. While $\norm{\wb(T)}$ monotonically increases for small norm initialization (SN), it decreases in the early phase of training and then increases for other initializations (MN, LN). 
Inspired from \citet{boursier2022gradient}, we conjecture that gradient flow dynamics has two phases ($\norm{\wb(T)}$ decreasing first, and increasing) for large norm initialization (LN) in general.

\section{Conclusion} \label{sec: conclusion}

In this paper, we analyzed the training dynamics of multi-layer single ReLU neurons with data following spherically symmetric distribution, by decomposing gradient flow $\wb(t)$ to magnitude $\norm{\wb(t)}$ and angle $\varphi(t)=\pi-\theta(t)$ components.
Investigating the decomposed dynamic system, we provided
upper and lower bounds of the magnitude $\norm{\wb(t)}$ and angle $\varphi(t)$ to describe gradient flow dynamics.
We reveal relation with initialization scale and convergence speed through our obtained bounds, and conclude that small norm initialization is prohibited in deep networks due to slow convergence.
Lastly, we 
extended the obtained bounds of magnitude and angle on gradient flow to gradient  descent, so that those bounds to be used in practice.
We also verified theoretical results by numerical experiments.

\paragraph{Limitation and future work}
Our work has several limitations.
First, we assumed spherically symmetric distribution on training data. Applying the technique introduced by \citet{wu2022learning}, it is expected to extend our results under weakened assumptions on training data distribution. 
Second, considering the bias term is required in our setting. Since the bias term is equivalent to adding $1$ component to the input dimension, this problem is also related with the distribution of training data.
Lastly, it is challenged to compute explicit form of the function $u(t;\varepsilon)$ in Theorem \ref{thm: norm of multi layer} for arbitrary $m>1$.

\newpage
{\small
\bibliography{aaai23.bib}
}

\newpage
\appendix
\section{Extended results to gradient  descent} \label{app: GD}
Here we provide gradient descent version of the theorems proposed in Section \ref{sec: GF}, via Theorem \ref{thm: relation between GF and GD}.
Proof of the following theorems is given in Appendix \ref{app: proof of app}.

\begin{theorem}[Gradient descent version of Theorem \ref{thm: norm of single neuron}] \label{thm: norm of single neuron GD version}
    Consider the dynamical system of a one-layer single ReLU neuron given in Proposition \ref{prop: ODE single neuron} trained by gradient descent with learning rate $\eta \ll 2$. Let $\varepsilon_0 := 1- \frac{\sin \varphi_0 - \varphi_0 \cos \varphi_0}{\pi}>0$ as in Theorem \ref{thm: norm of single neuron}. Then, upper and lower bounds of $\norm{\wb(T)}$ are given by
    \begin{align*}
        &(1-\varepsilon_0) \left(1 - (1-\frac12\eta)^T \right)  \norm{\wb^*} + \norm{\wb_0} (1-\frac12\eta)^T \\
        &< \norm{\wb(T)} 
        < \left(1 - (1-\frac12\eta)^T \right) \norm{\wb^*} + \norm{\wb_0}(1-\frac12\eta)^T .
    \end{align*}
    For the special case that the norm of initialization point is sufficiently small, i.e. $\norm{\wb_0}\approx 0$, the bounds are given by
    \begin{align*}
        & (1-\varepsilon_0) \left(1-(1-\frac12 \eta)^T\right) \norm{\wb^*} \\
        &\qquad < \norm{\wb(T)} < 
        \left(1-(1-\frac12 \eta)^T\right)\norm{\wb^*}.
    \end{align*}
\end{theorem}

\begin{theorem}[Gradient descent version of Theorem \ref{thm: angle of single neuron}]
    \label{thm: angle of single neuron GD version}
    Consider the dynamical system of a one-layer single ReLU neuron given in Proposition \ref{prop: ODE single neuron} trained by gradient descent with learning rate $\eta \ll \min\{ \frac{2\pi R}{\norm{\wb^*}\varphi_0}, \frac{2r}{3\norm{\wb^*}} \}$.
    Suppose there exist constants $r,R>0$ such that $r<\norm{\wb(T)}<R$ for all $T>0$. Then, $\varphi(T)$ is bounded by
    \begin{align*}
        \pi - &2 \cot \frac{\varphi_0}{2} \left(1 -\frac{\norm{\wb^*}\varphi_0}{2\pi R} \eta \right)^T
        < \; \varphi(T) \\
        & <  \pi - 2 \cot \frac{\varphi_0}{2} \left(1 -\frac{\norm{\wb^*}}{2r} \eta \right)^T \\
        &\qquad + \frac23 \cot^3 \frac{\varphi_0}{2} \left(1 -\frac{3\norm{\wb^*}}{2r} \eta \right)^T.
    \end{align*}
    For the special case that the norm of initialization point is sufficiently small, i.e. $\norm{\wb_0}\approx 0$, the bounds are given by
    \begin{align*}
        \pi - 2\cot \frac{\varphi_0}{2} \left( 1- \frac{\varphi_0}{2\pi} \eta \right)^T
        < \varphi(T) < \pi .
    \end{align*}
\end{theorem}

\begin{theorem}[Gradient descent version of Theorem \ref{thm: norm of multi layer}, when $m=1$] \label{thm: norm of two layer GD version}
    Consider the dynamical system of a two-layer single ReLU neuron in Proposition \ref{prop: ODE multi layer} with balanced initialization, trained by gradient descent with learning rate $\eta \ll \frac{1}{v^{*2}}$. Let $\varepsilon_0 = 1 - \frac{\sin \varphi_0 -\varphi_0 \cos \varphi_0 }{\pi}$.
    Then, magnitude $\norm{\wb(T)}=v(T)$ is bounded by
    \begin{align*}
        &v^* \sqrt{\frac{1-\varepsilon_0}{1 - \left(1 -  (1-\varepsilon_0) (\frac{v^*}{v_0})^2 \right) \left(1-(1-\varepsilon_0)v^{*2}\eta\right)^T}} \\
        &\hspace{1.7cm} < v(T) <
        v^* \sqrt{\frac{1}{1-(1-(\frac{v^*}{v_0})^2) \left(1-v^{*2}\eta\right)^T}} . 
    \end{align*}
\end{theorem}

\section{Discussion on data distribution} \label{app: uniform sphere}

In this section, we show that our work can be extended on any spherically symmetric data distribution.
First, we consider uniform distribution on sphere.
\begin{proposition}
    Suppose $\xb$ and $\zb$ follow the uniform distribution on the unit sphere and the standard Gaussian distribution, respectively.
    Let $\ub, \vb$ be two distinct unit vectors in $\Rd^d$. Then,
    \begin{align*}
        \Ed_{\xb} [\indicator{\xb^\top\ub >0} \xb\xb^\top ] &= \frac1d ~\Ed_{\zb} [\indicator{\zb^\top\ub >0} \zb\zb^\top ], \\
        \Ed_{\xb} [ \indicator{\ub^\top\xb>0}\indicator{\vb^\top\xb>0}\xb\xb^\top]
        &= \frac1d ~ \Ed_{\zb} [ \indicator{\ub^\top\zb>0}\indicator{\vb^\top\zb>0}\zb\zb^\top]
    \end{align*}
\end{proposition}
\begin{proof}
    Since the distribution has the spherical symmetry, the computation can be done in the same fashion as the isotropic Gaussian distribution. The only difference is the constant scale of $\frac1d$ by considering the surface measure\citep{10.2307/2690391}.
\end{proof}

By the above proposition, we can derive an analogous result of Lemma \ref{lem: two indicators} for uniform distribution on a unit sphere, by multiplying $\frac1d$.
Finally, since any spherically symmetric distribution is a mixture of uniform distributions on spheres \citep{Fourdrinier2018}, we get equivalent results to Lemma \ref{lem: two indicators} up to a constant multiple.

Now, we verify the uniqueness of $\wb^*$. Since the angle $\varphi(t)$ is defined by $\wb(t)$ and $\wb^*$, the uniqueness of $\wb^*$ is required to check. Fortunately, the following proposition guarantees the uniqueness if the data distribution has a full-rank covariance matrix. 

\begin{proposition} \label{prop: unique global minimum}
    Let $f_\wb$ and $f_{\wb^*}$ be one-layer single ReLU neurons. Suppose the covariance matrix $\Sigma:=\Ed[\xb\xb^\top]$ of data distribution $\xb\sim\Dc$ with zero mean has full rank.
    Then, the loss function 
    \begin{align} \label{eq: loss function general}
        L(\wb) := \frac12 \Ed_{\xb\sim\Dc} [(f_{\wb}(\xb)-f_{\wb^*}(\xb))^2]
    \end{align}
    has the unique global minimum $\wb^*$.
\end{proposition}
\begin{proof}
    Trivially, $\wb^*$ is a global minimum of \eqref{eq: loss function general} since $L \ge 0$ and $L(\wb^*)=0$. Now suppose $L(\wb_1)=L(\wb_2)=0$. Then, for all $\xb$, we get $\xb^\top\wb_1 = \xb^\top\wb_2$. In other words, $\xb^\top(\wb_1 - \wb_2)=0$. This implies $\xb\xb^\top(\wb_1 - \wb_2)=\zerob$. Finally, taking expectation on distribution, we get
    \begin{align*}
        \zerob = \Ed[\xb\xb^\top(\wb_1 - \wb_2)]
        = \Ed[\xb\xb^\top](\wb_1 - \wb_2),
    \end{align*}
    and conclude that $\wb_1 = \wb_2$.
\end{proof}

\section{Proof of Lemmas, Propositions, and Theorems}
Here we provide the complete proofs of lemmas, propositions, and theorems in the paper.

\subsection{Proofs for Section \ref{sec: GF}} \label{app: proof of sec one layer}
\newcommand{\s}{\boldsymbol{s}}
\begin{proof}[Proof of Lemma \ref{lem: two indicators}]
    To compute the matrix, let $\pb, \qb$ be two unit vectors and consider $\pb^\top\Ed[ \indicator{\ub^\top\xb>0}\indicator{\vb^\top\xb>0}\xb\xb^\top] \qb$.
    Since $\xb$ is a spherical Gaussian, $\pb^\top\xb$ and $\qb^\top\xb$ are independent if $\pb \perp \qb$.
    When $\pb = \qb = \ub$, by symmetry, 
    \begin{align*}
        \Ed[ \indicator{\ub^\top\xb>0}(\ub^\top\xb)^2] = \frac12 \Ed[ (\ub^\top\xb)^2] = \frac12 .
    \end{align*}
    When $\qb \perp \pb = \ub$, 
    \begin{align*}
        &\Ed[ \indicator{\ub^\top\xb>0}(\ub^\top\xb)(\qb^\top\xb)] \\
        &\qquad = \Ed[ \indicator{\ub^\top\xb>0}(\ub^\top\xb)]\Ed[\qb^\top\xb] \\
        &\qquad = 0.
    \end{align*}
    When $\pb = \qb \perp \ub $, 
    $$\Ed[ \indicator{\ub^\top\xb>0}(\pb^\top\xb)^2] = \Ed[\indicator{\ub^\top\xb>0}]\Ed [(\pb^\top\xb)^2] = \frac12 .$$
    When $\pb$, $\qb$ and $\ub$ are mutually orthogonal,
    \begin{align*}
        &\Ed[ \indicator{\ub^\top\xb>0}(\pb^\top\xb)(\qb^\top\xb)] \\
        &\qquad = \Ed[ \indicator{\ub^\top\xb>0}]\Ed[(\pb^\top\xb)]\Ed[\qb^\top\xb] \\
        &\qquad = 0.
    \end{align*}
    To sum up, since all eigenvalues are eqaul to $\frac12$, $$\Ed[ \indicator{\ub^\top\xb>0}\xb\xb^\top] = \frac12 \Ib.$$
    
    In similar fashion, we first predict the eigenvectors of the matrix, and attain eigenvalues by
    computing $\pb^\top\Ed[ \indicator{\ub^\top\xb>0}\indicator{\vb^\top\xb>0}\xb\xb^\top] \qb$. Indeed, we define $\rb:= \frac{u+v}{2\cos(\theta/2)}$ and $\s:= \frac{u-v}{2\sin(\theta/2)}$ and verify that they are the eigenvectors.
    
    To compute expectation, we decompose the whole space as $\Rd^d = W \oplus W^{\perp}$, where $W:= \text{span}(\ub,\vb)$.
    If $f: \Rd^d \rightarrow \Rd$ is decomposed as $f(\xb_W + \xb_{W^{\perp}}) = f_W(\xb_W)f_{W^{\perp}}(\xb_{W^{\perp}})$ for all $\xb_W \in W$ and $\xb_{W^{\perp}}\in{W^{\perp}}$, we can get simpler form of the expectation containing indicators:
    \begin{align*}
        \Ed[&\indicator{\ub^\top\xb>0}\indicator{\vb^\top\xb>0}f(\xb)]\\
        =& \int_{\{\xb \in \Rd^d: \ub^\top\xb>0, \vb^\top\xb>0\}} f(\xb) (2\pi)^{-\frac12 d} e^{-\frac12 \norm{\xb}^2} d\xb \\
        =& \int_{W\cap \{\yb \in \Rd^d: \ub^\top\yb>0, \vb^\top\yb>0\}} f_W(\yb) (2\pi)^{-1} e^{-\frac12 \norm{\yb}^2} d\yb \\
        &\times \int_{W^\perp} f_{W^{\perp}}(\zb) (2\pi)^{-\frac12 d+1} e^{-\frac12 \norm{\zb}^2} d\zb.
    \end{align*}
    $\pb=\qb=\rb$ :
    \begin{align*}
        &\int_{\{\xb \in \Rd^d: \ub^\top\xb>0, \vb^\top\xb>0\}} (\rb^\top\xb)^2 (2\pi)^{-\frac12 d} e^{-\frac12 \norm{\xb}^2} d\xb \\
        &= \int_{W\cap \{\yb \in \Rd^d: \ub^\top\yb>0, \vb^\top\yb>0\}} (\rb^\top\yb)^2 (2\pi)^{-1} e^{-\frac12 \norm{\yb}^2} d\yb\\
        &\qquad \times\int_{W^\perp} (2\pi)^{-\frac12 d+1} e^{-\frac12 \norm{\zb}^2} d\zb\\
        &= \iint_{(0,\infty)\times(-\frac{\pi-\theta}{2},\frac{\pi-\theta}{2})} r^2 \cos^2\alpha (2\pi)^{-1} e^{-\frac12 r^2} r dr d\alpha \\
        &= \frac{\pi - \theta + \sin\theta}{2\pi}
    \end{align*}
    In the last line, we introduce polar coordinates $(r, \alpha)$ of $W$ which regards $\rb$ as $\alpha = 0$. We repeatedly employ the same polar coordinates below.
    
    $\pb = \qb = \s$:
    \begin{align*}
        &\int_{\{\xb \in \Rd^d: \ub^\top\xb>0, \vb^\top\xb>0\}} (\s^\top\xb)^2 (2\pi)^{-\frac12 d} e^{-\frac12 \norm{\xb}^2} d\xb\\
        &= \iint_{(0,\infty)\times(-\frac{\pi-\theta}{2},\frac{\pi-\theta}{2})} r^2 \sin^2\alpha (2\pi)^{-1} e^{-\frac12 r^2} r dr d\alpha\\
        &= \frac{\pi-\theta-\sin\theta}{2\pi}
    \end{align*}
    $\pb = \rb, \qb = \s$:
    \begin{align*}
        &\int_{\{\xb \in \Rd^d: \ub^\top\xb>0, \vb^\top\xb>0\}} (\rb^\top\xb)(\s^\top\xb) (2\pi)^{-\frac12 d} e^{-\frac12 \norm{\xb}^2} d\xb\\
        &= \iint_{(0,\infty)\times(-\frac{\pi-\theta}{2},\frac{\pi-\theta}{2})} r^2 \cos\alpha \sin\alpha (2\pi)^{-1} e^{-\frac12 r^2} r dr d\alpha \\
        &= 0
    \end{align*}
    $\pb \in W$ and $\qb \in W^{\perp}$:
    \begin{align*}
        &\int_{\{\xb \in \Rd^d: \ub^\top\xb>0, \vb^\top\xb>0\}} (\pb^\top\xb)(\qb^\top\xb) (2\pi)^{-\frac12 d} e^{-\frac12 \norm{\xb}^2} d\xb\\
        &= \int_{W\cap \{\yb \in \Rd^d: \ub^\top\yb >0, \vb^\top\yb>0\}} (\pb^\top\yb) (2\pi)^{-1} e^{-\frac12 \norm{\yb}^2} d\yb \\
        &\quad \times \int_{W^\perp} (\qb^\top\zb) (2\pi)^{-\frac12 d+1} e^{-\frac12 \norm{\zb}^2} d\zb \\
        &= 0
    \end{align*}
    $\pb = \qb \in W^{\perp}$:
    \begin{align*}
        &\int_{\{\xb \in \Rd^d: \ub^\top\xb>0, \vb^\top\xb>0\}} (\pb^\top\xb)^2 (2\pi)^{-\frac12 d} e^{-\frac12 \norm{\xb}^2} d\xb\\
        &= \int_{W\cap \{\yb \in \Rd^d: \ub^\top\yb >0, \vb^\top\yb>0\}} (2\pi)^{-1} e^{-\frac12 \norm{\yb}^2} d\yb \\
        &\quad \times \int_{W^\perp} (\qb^\top\zb)^2 (2\pi)^{-\frac12 d+1} e^{-\frac12 \norm{\zb}^2} d\zb \\
        &= \frac{\pi-\theta}{2\pi}
    \end{align*}
    We used calculation of integrals : $\int_0^\infty r^3 e^{-\frac12 r^2} dr = 2$ and 
    \begin{align*}
        \int \cos^2 \alpha d\alpha &= \frac{1}{2}\alpha + \frac{1}{4}\sin 2\alpha, \\
        \int \sin^2 \alpha d\alpha &= \frac{1}{2}\alpha - \frac{1}{4}\sin 2\alpha,
    \end{align*}
    and the mean and the variance of Gaussian variable,
    \begin{align*}
        \int_{W^\perp} (\qb^\top\zb) (2\pi)^{-\frac12 d+1} e^{-\frac12 \norm{\zb}^2} d\zb = 0, \\
        \int_{W^\perp} (\qb^\top\zb)^2 (2\pi)^{-\frac12 d+1} e^{-\frac12 \norm{\zb}^2} d\zb = 1.
    \end{align*}
    To sum up, by eigendecomposition of a matrix,
    \begin{align*}
        &\Ed[ \indicator{\ub^\top\xb>0}\indicator{\vb^\top\xb>0}\xb\xb^\top]\\
        &= (\frac{\pi-\theta+\sin\theta}{2\pi}) \rb\rb^\top + (\frac{\pi-\theta-\sin\theta}{2\pi}) \s\s^\top \\
        & \quad + (\frac{\pi-\theta}{2\pi})(\Ib - \rb\rb^\top - \s\s^\top) \\
        &= \frac{\pi-\theta}{2\pi} \Ib + \frac{\sin \theta}{2\pi} (\rb\rb^\top - \s\s^\top) \\
        &= \frac{\pi-\theta}{2\pi} \Ib + \frac{\sin \theta}{2\pi} 
        \bigg( \frac{(\ub+\vb)(\ub+\vb)^\top}{4\cos^2 \frac\theta2} \\
        &\hspace{3.4cm} -\frac{(\ub-\vb)(\ub-\vb)^\top}{4\sin^2 \frac\theta2}
        \bigg) \\
        &= \frac{\pi-\theta}{2\pi} \Ib + \frac{\sin \theta}{2\pi} 
        \bigg( \frac{\ub\ub^\top + \vb\vb^\top + \ub\vb^\top + \vb\ub^\top}{2(1+\cos\theta)} \\
        & \qquad \qquad \qquad \qquad \quad - \frac{\ub\ub^\top + \vb\vb^\top - \ub\vb^\top - \vb\ub^\top}{2(1-\cos\theta)} \bigg) \\
        &= \frac{\pi-\theta}{2\pi} \Ib + \frac{\sin \theta}{2\pi} 
        \bigg(\frac12 
        (\frac{1}{1+\cos\theta} - \frac{1}{1-\cos\theta})
        ( \ub\ub^\top + \vb\vb^\top ) \\
        & \qquad \qquad \qquad + \frac12 (\frac{1}{1+\cos\theta} + \frac{1}{1-\cos\theta})
        ( \ub\vb^\top + \vb\ub^\top )
        \bigg) \\
        &= \frac{\pi-\theta}{2\pi} \Ib + \frac{\sin \theta}{2\pi} 
        \bigg( -\frac{\cos\theta}{\sin^2\theta} ( \ub\ub^\top + \vb\vb^\top ) \\ 
        &\qquad \qquad \qquad \qquad \quad  + \frac{1}{\sin^2\theta} ( \ub\vb^\top + \vb\ub^\top )
        \bigg) \\
        &= \frac12 \left( 1- \frac\theta\pi \right) \Ib + \frac{1}{2\pi\sin\theta} 
        \big( -\cos\theta(\ub\ub^\top + \vb\vb^\top ) \\
        &\hspace{4.3cm} + ( \ub\vb^\top + \vb\ub^\top ) \big).
    \end{align*}
\end{proof}

\begin{proof}[Proof of Proposition \ref{prop: ODE single neuron}]
    Using Lemma \ref{lem: two indicators}, from \eqref{eq: gradient flow single neuron},
    \begin{align*}
        \dot \wb 
        &= -\Ed [\indicator{\xb^\top\wb >0} \sigma(\wb^\top\xb)\xb ] + \Ed[\indicator{\xb^\top\wb >0} \sigma(\wb^{\top}\xb)\xb] \\
        &= -\Ed [\indicator{\xb^\top\wb >0} \xb\xb^\top ]\wb + \Ed[\indicator{\xb^\top\wb >0, \;  \xb^\top\wb^*>0} \xb\xb^\top]\wb^* \\
        &= -\frac12 \wb + \bigg[ 
        \frac12 (1-\frac\theta\pi) \Ib + \frac{1}{2\pi\sin\theta} \\
        &\qquad \times[ -\cos\theta (\frac{\wb\wb^\top}{\norm{\wb}^2} + \frac{\wb^*\wb^{\top}}{\norm{\wb^*}^2})
        + \frac{\wb^*\wb^\top+\wb\wb^{\top}}{\norm{\wb}\norm{\wb^*}} ]
        \bigg] \wb^*\\
        &= -\frac12 \wb + \frac12 (1-\frac\theta\pi)\wb^*
        + \frac{1}{2\pi\sin\theta}\Big[ \\
        &\qquad -\cos\theta (\frac{\norm{\wb}\norm{\wb^*}\cos\theta}{\norm{\wb}^2}\wb
         + \wb^*) +\wb^*\cos\theta + \frac{\norm{\wb^*}}{\norm{\wb}}\wb \Big] \\
        &= -\frac12 \wb + \frac12 (1-\frac\theta\pi)\wb^* + \frac{1}{2\pi\sin\theta}(1-\cos^2\theta)\frac{\norm{\wb^*}}{\norm{\wb}} \wb \\
        &= -\frac12 \wb + \frac12 (1-\frac{\theta}{\pi})\wb^* + \frac{\sin \theta}{2\pi} \frac{\norm{\wb^*}}{\norm{\wb}}\wb \\
        &= -\frac12 \wb + \frac{\varphi}{2\pi}\wb^* + \frac{\sin\varphi}{2\pi} \frac{\norm{\wb^*}}{\norm{\wb}}\wb .
    \end{align*}
    Moreover, from $\norm{\wb}\frac{d}{dt} \norm{\wb} = \frac12 \frac{d}{dt} \norm{\wb}^2 = \frac12 \frac{d}{dt} (\wb^\top\wb) = \wb^\top \dot\wb $, we get $\frac{d}{dt}\norm{\wb} = \frac{1}{\norm{\wb}} \wb^\top\dot\wb$. Then, 
    \begin{align*} 
        \frac{d}{dt}\norm{\wb} &= \frac{1}{\norm{\wb}} \wb^\top\dot\wb \\
        &= \frac{1}{\norm{\wb}} \wb^\top \left( -\frac12 \wb + \frac{\varphi}{2\pi}\wb^* + \frac{\sin\varphi}{2\pi} \frac{\norm{\wb^*}}{\norm{\wb}}\wb \right) \\
        &= -\frac12 \norm{\wb} + \frac{\varphi}{2\pi} \norm{\wb^*}\cos\theta +\frac{1}{2\pi}\sin\varphi \norm{\wb^*} \\
        &= -\frac12 \norm{\wb} + \frac12 \norm{\wb^*}  \frac{\sin\varphi - \varphi \cos \varphi}{\pi} .
    \end{align*}
    Similarly, differentiating both sides of $\cos \theta = \frac{\wb^\top\wb^*}{  \norm{\wb} \norm{\wb^*}}$,
    \begin{align*}
        &-\sin \theta \frac{d\theta}{dt} \\
        &= (\frac{\wb^*}{\norm{\wb^*}})^\top \frac{\dot\wb\norm{\wb} - \wb \frac{d}{dt}\norm{\wb}}{\norm{\wb}^2} \\
        &= (\frac{\wb^*}{\norm{\wb^*}})^\top \frac{1}{\norm{\wb}^2} \bigg[
        -\frac12 \norm{\wb}\wb +\frac12 (1-\frac{\theta}{\pi}) \norm{\wb}\wb^* \\
        &\hspace{3cm} + \frac{\sin \theta}{2\pi}\norm{\wb^*}\wb  \\
        &\quad -\wb \left(-\frac12 \norm{\wb} + \frac12 \norm{\wb^*} [(1-\frac{\theta}{\pi})\cos \theta +\frac{1}{\pi}\sin \theta ]\right) \bigg] \\
        &= (\frac{\wb^*}{\norm{\wb^*}})^\top \frac{1}{\norm{\wb}^2}
        \frac12 (1-\frac{\theta}{\pi})[ \norm{\wb}\wb^* - \norm{\wb^*}\cos \theta \wb] \\
        &= \frac12 \left( 1-\frac{\theta}{\pi} \right) \frac{1}{\norm{\wb}^2\norm{\wb^*}}
        \Big[ \norm{\wb}\norm{\wb^*}^2 \\
        &\hspace{4.4cm} -\norm{\wb}\norm{\wb^*}^2\cos^2\theta \Big] \\
        &= \frac12 \left( 1-\frac{\theta}{\pi}\right) \frac{\norm{\wb^*}}{\norm{\wb}} \sin^2 \theta.
    \end{align*}
    Dividing both sides by $-\sin\theta$, we get $\frac{d\varphi}{dt} = -\frac{d\theta}{dt} = \frac{\varphi}{2\pi} \frac{\norm{\wb^*}}{\norm{\wb}}\sin\varphi$.
\end{proof}

\begin{proof}[Proof of Theorem \ref{thm: norm of single neuron}]
    Let $\varepsilon(t) = 1-\frac{\sin \varphi(t) - \varphi(t) \cos \varphi(t)}{\pi}$. Then \eqref{eq: norm w} gives
    \begin{align*}
        \frac{d}{dt}\norm{\wb} = -\frac12 \norm{\wb} + \frac12 \norm{\wb^*}(1-\varepsilon).
    \end{align*}
    Since $\varphi(t)$ monotonically increases, $0<\frac{\sin \varphi_0 - \varphi_0 \cos \varphi_0}{\pi}<1$ also monotonically increases and we get
    \begin{align*}
        \frac12 \norm{\wb^*}(1-\varepsilon_0) <
        \frac{d}{dt}\norm{\wb} + \frac12 \norm{\wb} <
        \frac12 \norm{\wb^*}.
    \end{align*}
    Now we use Gronwall comparison. Using $\frac{d}{dt}[e^{\frac12 t}\norm{\wb} ]= e^{\frac12 t} (\frac{d}{dt} \norm{\wb} + \frac12 \norm{\wb}) $, it deforms to
    \begin{align*}
        \frac12 \norm{\wb^*}(1-\varepsilon_0) e^{\frac12 t} <
        \frac{d}{dt}[e^{\frac12 t}\norm{\wb} ] <
        \frac12 \norm{\wb^*}e^{\frac12 t}.
    \end{align*}
    Integrating from $0$ to $t$, we get
    \begin{align*}
        (1-&\varepsilon_0)(1-e^{-\frac12 t})\norm{\wb^*} + \norm{\wb_0}e^{-\frac12 t} \\
        &< \norm{\wb(t)} <
        (1-e^{-\frac12 t})\norm{\wb^*} + \norm{\wb_0}e^{-\frac12 t}.
    \end{align*}
    Taking $\norm{\wb_0}\rightarrow 0$, we get the last statement of the theorem.
\end{proof}

\begin{proof}[Proof of Theorem \ref{thm: angle of single neuron}]
    Note that $\varphi(t)$ strictly increases by Theorem \ref{thm: global convergence one layer}, thus $\varphi_0  < \varphi(t) < \pi$ . From \eqref{eq: angle phi}, we get $\csc \varphi\frac{d\varphi}{dt} 
        = \frac12 \frac{\norm{\wb^*}}{\norm{\wb}}\frac{\varphi}{\pi}$ and
    \begin{align*}
        \frac12 \frac{\norm{\wb^*}}{R} \frac{\varphi_0}{\pi}
        < \csc \varphi\frac{d\varphi}{dt}
        < \frac12 \frac{\norm{\wb^*}}{r}
    \end{align*}
    
    Therefore, $\frac{\norm{\wb^*}}{2R}\frac{\varphi_0}{\pi} dt < \csc \varphi \; d\varphi < \frac{\norm{\wb^*}}{2r} dt $. Integrating both sides from initial state to time $t$ (Grownall comparison), we get
    \begin{align*}
        \frac{\norm{\wb^*}}{2R} \frac{\varphi_0}{\pi} t
        < \log \frac{\tan \frac{\varphi}{2}}{\tan \frac{\varphi_0}{2}}
        < \frac{\norm{\wb^*}}{2r}t,
    \end{align*}
    which implies 
    \begin{align*}
        \arctan \left( \tan \frac{\varphi_0}{2} \;  e^{\frac{\norm{\wb^*}}{2R}\frac{\varphi_0}{\pi} t} \right) 
        &< \frac{\varphi}{2} \\
        &< \arctan \left( \tan \frac{\varphi_0}{2} \;  e^{\frac{\norm{\wb^*}}{2r}t} \right).
    \end{align*}
    Using a well-known inequality $\frac\pi2 -\frac1x \le \arctan x \le \frac\pi2 - \frac1x + \frac{1}{3x^3}$, we conclude
    \begin{align*}
        \frac\pi2 - &\cot \frac{\varphi_0}{2} \;  e^{-\frac{\norm{\wb^*}}{2R}\frac{\varphi_0}{\pi} t}
        < \frac\varphi2 \\
        &< \frac\pi2 - \cot \frac{\varphi_0}{2} \;  e^{-\frac{\norm{\wb^*}}{2r}t} + \frac13 \cot^3 \frac{\varphi_0}{2} \; e^{-\frac{3\norm{\wb^*}}{2r}t}
    \end{align*}
    which completes the proof.
    
\end{proof}

\begin{proposition} \label{prop: random initialization angle}
    Let $\ub, \vb \in \Rd^d$ be independent samples chosen from spherical Gaussian. Then, for arbitrary $\varepsilon>0$,
    \begin{align*}
        P\left( \frac{\ub^\top \vb}{\norm{\ub}\norm{\vb}} < \varepsilon\right) \ge 1 - 2 \exp ({-\frac12 {d\varepsilon^2}}).
    \end{align*}
\end{proposition}
\begin{proof}[Proof of Proposition \ref{prop: random initialization angle}]
The distribution of $\frac{u}{\norm{u}}, (\frac{v}{\norm{v}})$ are equal to the uniform distribution on the unit sphere. Then, we directly apply a result about measure on a high dimensional sphere\cite[Lemma 2.2]{ball1997elementary}.
$$
P\left( (\frac{u}{\norm{u}})^\top \frac{v}{\norm{v}} > \varepsilon \right) \le (1-\varepsilon^2)^{\frac d2} \le \exp(-\frac{d\epsilon^2}{2}).
$$
\end{proof}

\subsection{Proofs for Section \ref{sec: multi layer ReLU neuron}} \label{app: proof of sec multi layer}

\begin{proof}[Proof of Proposition \ref{prop: equivalent to linear}]
    Note that a multi-layer single ReLU neuron could be replaced by $f(\xb) = v_1 \sigma(v_2) \sigma(v_3) \cdots \sigma(v_m) \sigma(\wb^\top\xb)$, since the output is equivalent. Then, the gradient of $v_k$ is given by
    \begin{align} 
        \frac{\partial L}{\partial v_k} = \Ed \Big[(f(\xb)-f^*(\xb))&v_1 \sigma(v_2) \cdots \sigma(v_{k-1})\indicator{v_k>0} \notag \\
        &\times\sigma(v_{k+1}) \cdots\sigma(v_m)\sigma(\wb^\top\xb)\Big]. \label{eq: gradient}
    \end{align}
    Suppose there exists $v_k\le0$ for some $2\le k \le m$. Then, by ReLU activation $\sigma$, both the network output and gradient are vanished by \eqref{eq: gradient}. Therefore, the gradient flow is stuck, and the sign of all $v_k$'s are obviously invariant.
    
    Now, we consider the situation that all $v_k$ ($2\le k\le m$) are positive. To show that the sign of $v_k$'s are invariant, it is enough to show that $\frac{dv_k}{dt}\Big|_{v_k=0}>0$ for all $2\le k\le m$. To check it,
    \begin{align*}
        \frac{dv_k}{dt}\Big|_{v_k=0} &= -\frac{\partial L}{\partial v_k}\Big|_{v_k=0} \\
        &= -\Ed \Big[ (f(\xb)-f^*(\xb))v_1 \sigma(v_2) \cdots \sigma(v_{k-1}) \\
        &\qquad \times\indicator{v_k>0}\sigma(v_{k+1}) \cdots\sigma(v_m)\sigma(\wb^\top\xb) \Big]\Big|_{v_k=0} \\
        &= v_1 v_2 \cdots v_{k-1} v_{k+1} \cdots v_m \Ed \Big[ f^*(\xb)\sigma(\wb^\top\xb) \Big] \\
        &> 0.
    \end{align*}
\end{proof}

\begin{proof}[Proof of Proposition \ref{prop: ODE multi layer}]
    The loss function \eqref{eq: loss function} is given by
    \begin{align*}
        L(\wb, v_1, \cdots, v_m) 
        &= \frac12 \Ed\left[\left(
        (\prod_{k=1}^m v_k)  \sigma(\wb^\top\xb) - (\prod_{k=1}^m v_k^*) \sigma(\wb^{*\top}\xb)
        \right)^2\right].
    \end{align*}
    From balanced condition, let $v(t) := v_1(t) = v_2(t) = \cdots =v_m(t)=\norm{\wb(t)}>0$ and $v^* := v_1^*= v_2^* = \cdots = v_m^*$. Then gradients are given by
    \begin{align*}
        \frac{\partial L}{\partial \wb}
        &= \Ed \Bigg[ 
        \left( (\prod_{k=1}^m v_k)  \sigma(\wb^\top\xb) - (\prod_{k=1}^m v_k^*) \sigma(\wb^{*\top}\xb) \right) \\
        &\qquad \quad \times (\prod_{k=1}^m v_k)\indicator{\wb^\top\xb>0}\xb
        \Bigg] \\
        &= (\prod_{k=1}^m v_k)^2 \Ed[\indicator{\wb^\top\xb>0} \xb\xb^\top ]\wb \\
        &\quad - (\prod_{k=1}^m v_k)(\prod_{k=1}^m v_k^*) \Ed[\indicator{\wb^{*\top}\xb>0}\indicator{\wb^{\top}\xb>0} \xb\xb^\top ]\wb^* \\
        &= v^{2m} \Ed[\indicator{\wb^\top\xb>0} \xb\xb^\top ]\wb \\
        &\quad - v^m v^{*m} \Ed[\indicator{\wb^{*\top}\xb>0}\indicator{\wb^{\top}\xb>0} \xb\xb^\top ]\wb^* \\
        &= \frac12 v^{2m} \wb - v^m v^{*m} \frac{1}{2\pi}(\varphi \wb^* + \frac{v^*}{v} \sin \varphi \;\wb), \\
        \frac{\partial L}{\partial v_1 } &= \Ed \Bigg[ 
        \left( (\prod_{k=1}^m v_k)  \sigma(\wb^\top\xb) - (\prod_{k=1}^m v_k^*) \sigma(\wb^{*\top}\xb) \right) \\
        &\qquad \quad \times \frac{(\prod_{k=1}^m v_k)}{v_1} \sigma(\wb^\top\xb) \Bigg] \\
        &= \frac{(\prod_{k=1}^m v_k)^2}{v_1} \wb^\top \Ed[\indicator{\wb^\top\xb>0} \xb\xb^\top ]\wb \\
        &\quad - \frac{(\prod_{k=1}^m v_k) (\prod_{k=1}^m v_k^*)}{v_1} \\
        &\quad \times \wb^\top\Ed[\indicator{\wb^{*\top}\xb>0}\indicator{\wb^{\top}\xb>0} \xb\xb^\top ]\wb^* \\
        &= \frac12 v^{2m-1} \norm{\wb}^2 - v^{m-1}v^{*m} \frac{\sin \varphi - \varphi \cos \varphi}{2\pi} vv^* \\
        &= \frac12 v^{2m+1} - v^m v^{*m+1} \frac{\sin \varphi - \varphi \cos \varphi}{2\pi}.
    \end{align*}
    Therefore, 
    \begin{align*}
        \frac{d}{dt}\wb &= -\frac12 v^{2m} \wb + \frac{1}{2\pi} v^m v^{*m} (\varphi \wb^* + \frac{v^*}{v} \sin \varphi \;\wb) \\
        \frac{dv}{dt} &= -\frac12 v^{2m+1} + \frac12 v^m v^{*m+1} \frac{\sin \varphi - \varphi \cos \varphi}{\pi}.
    \end{align*}
    For derivative of $\varphi(t)$, we start with $\cos \varphi = - \frac{\wb^\top\wb^*}{vv^*}$. Differentiating both sides,
    \begin{align*}
        \sin& \varphi \frac{d\varphi}{dt} \\
        &= (\frac{\wb^*}{v^*})^\top \frac{d}{dt} \left(\frac{\wb}{v}\right) \\
        &= (\frac{\wb^*}{v^*})^\top (\frac{\dot\wb v - \dot v \wb }{v^2}) \\
        &= (\frac{\wb^*}{v^*})^\top \frac{1}{v^2} 
        \bigg[\\
        &\quad\left(-\frac12 v^{2m} \wb + \frac{1}{2\pi} v^m v^{*m} (\varphi \wb^* + \frac{v^*}{v} \sin \varphi \;\wb) \right) v \\
        &\quad -\left( -\frac12 v^{2m+1} + \frac12 v^m v^{*m+1} \frac{\sin \varphi - \varphi \cos \varphi}{\pi} \right) \wb
        \bigg] \\
        &= \frac{1}{v^2v^*}\wb^{\top} \bigg[ 
        \frac{1}{2\pi} v^{m+1} v^{*m} (\varphi \wb^* + \frac{v^*}{v} \sin \varphi \wb) \\
        &\hspace{2cm} - \frac12 v^m v^{*m+1} \frac{\sin \varphi - \varphi \cos \varphi}{\pi}\wb \bigg] \\
        &= \frac{1}{v^2v^*} \bigg[ 
        \frac{1}{2\pi} v^{m+1} v^{*m} (\varphi v^{*2} - \frac{v^*}{v} \sin \varphi \cos \varphi \; vv^*) \\
        &\hspace{1.4cm}+ \frac12 v^m v^{*m+1} \frac{\sin \varphi - \varphi \cos \varphi}{\pi}vv^* \cos \varphi \bigg] \\
        &= \frac{1}{2\pi} v^{m-1}v^{*m+1} \varphi (1-\cos^2 \varphi) \\
        &= v^{m-1}v^{*m+1} \frac{\varphi \sin^2 \varphi }{2\pi}.
    \end{align*}
    Dividing both sides by $\sin \varphi$ completes the proof.
\end{proof}

\begin{proof}[Proof of Theorem \ref{thm: global convergence of multi layer}]
    Proof is almost same with the proof of Theorem \ref{thm: global convergence one layer}. First, from \eqref{eq: angle deep}, we know $\frac{d\varphi}{dt}>0$. Thus $\varphi(t)$ strictly increases, and converges to some value $\tilde \varphi$. By the same argument in the proof of Theorem \ref{thm: global convergence one layer}, we conclude $\varphi(\infty)=\pi$. 
    Now, it is remained to show that $v(\infty) = v^*$. We know $\varphi(\infty)=\pi$ and $\frac{dv}{dt}\big|_{t=\infty} =0$, thus \eqref{eq: norm v} implies $v(\infty)^{m+1} - v^{*m+1}=0$. Therefore, we conclude $v(\infty)=v^*$, which completes the proof.
\end{proof}

\begin{lemma} \label{lem: ODE-ineq}
    Consider two ODEs :
    \begin{gather*}
        \dot{u} = f(u) \\
        \dot{v} = g(v).
    \end{gather*}
    If $u(t_0)\le v(t_0)$ and $f(x)\le g(x)$ for all $x$, then $u(t) \le v(t)$ for all $t\ge t_0$.
\end{lemma}
\begin{proof}[Proof of Lemma \ref{lem: ODE-ineq}]
    We prove this by contradiction. Suppose $u(t) > v(t)$ for some $t$. Then, $\Tc:= \{t \ge t_0 : u(t) > v(t) \}$ is nonempty and we can take its infimum $t' = \inf \Tc$.
    Then, by continuity, $u(t') = v(t')$.
    It follows that $\dot{u}(t') = f(u(t')) \le g(v(t')) = \dot{v}(t')$.
    Hence, there exists small $\varepsilon>0$ such that $u(t) \le v(t)$ for $t \in [t', t'+\varepsilon)$, which contradicts to $t' = \inf \Tc$.
\end{proof}

\begin{proof}[Proof of Theorem \ref{thm: norm of multi layer}]
    For $\varepsilon > 0$, consider the following differential equation:
    \begin{align} \label{eq: hypergeometric}
        \frac{dv}{dt} = -\frac12 v^m (v^{m+1} + v^{*m+1}(1-\varepsilon)), \qquad v(0)=v_0.
    \end{align}
    This is solved by separable of variables. Integrating both sides of  $\frac{1}{v^m (v^{m+1} + v^{*m+1}(1-\varepsilon))} dv = -\frac12 dt$, we get
    \begin{align*}
        &\frac{v^{1-m}}{(m-1)v^{*m+1}(1-\varepsilon)} \ _2F_1(1, \frac{1-m}{m+1}; \frac{2}{m+1}; \frac{v^{m+1}}{v^{*m+1}(1-\varepsilon)}) \\
        &= -\frac12 t + C
    \end{align*}
    where $\ _2F_1(a,b;c;z)$ denotes the Gaussian hypergeometric function and $C$ is the constant of integration.
    Substituting the initial condition($t=0$), we get 
    \begin{align*}
        &C = \\
        &\frac{v_0^{1-m}}{(m-1)v^{*m+1}(1-\varepsilon_0)} \ _2F_1(1, \frac{1-m}{m+1}; \frac{2}{m+1}; \frac{v_0^{m+1}}{v^{*m+1}(1-\varepsilon_0)}).
    \end{align*}
    Therefore, we conclude that the function $u(t;\varepsilon)$ defined in Theorem \ref{thm: norm of multi layer} is the solution of \eqref{eq: hypergeometric}.
    
    Now, we recall \eqref{eq: norm v}. From Theorem \ref{thm: global convergence of multi layer}, we know $\varphi(t)$ strictly increases, and thus $\frac{\sin \varphi - \varphi \cos \varphi}{\pi}$ also strictly increases. Defining $\varepsilon(t):= 1- \frac{\sin \varphi(t) - \varphi(t) \cos \varphi(t)}{\pi}$, we conclude that $\varepsilon(t)$ decreases from $\varepsilon_0$ to $0$. Therefore, combined with \eqref{eq: norm v}, we get
    \begin{align*}
        -\frac12 v^m &(v^{m+1} + v^{*m+1}(1-\varepsilon_0)) \\
        &< \frac{dv}{dt} \\
        &= -\frac12 v^m (v^{m+1} + v^{*m+1}(1-\varepsilon))\\
        &< -\frac12 v^m (v^{m+1} + v^{*m+1}).
    \end{align*}
    Finally, by Lemma \ref{lem: ODE-ineq}, $v(t)$ is bounded by the solution of differential equations, which are given by $u(t; \varepsilon_0)$ and $u(t;0)$. This proves the first part.

    For a special case $m=1$, we can directly solve \eqref{eq: hypergeometric}.
    By the same argument above, it is enough to obtain solutions of the following two ODEs :
    \begin{align*}
        \dot{v_1} &= -\frac12 v_1 (v_1^2 -v^{*2} + v^{*2} \varepsilon_0 ), \\
        \dot{v_2} &= -\frac12 v_2 (v_2^2 -v^{*2}).
    \end{align*}
    These ODEs can be easily solved by separation of variables. The solution is given by
    \begin{align*}
        v_1 (t) &= v^* \sqrt{\frac{1-\varepsilon_0}{1 - e^{-(1-\varepsilon_0)v^{*2}t} (1 - (1-\varepsilon_0)(\frac{v^*}{v_1(0)})^2) }}, \\
        v_2 (t) &=
        v^* \sqrt{\frac{1}{1-e^{-v^{*2}t}(1-(\frac{v^*}{v_2(0)})^2)}}.
    \end{align*}
    
    By Lemma \ref{lem: ODE-ineq}, we have $v_1(t) < v(t) < v_2(t)$ for all $t \ge 0$.
    i.e.,
    \begin{align*}
        &v^* \sqrt{\frac{1-\varepsilon_0}{1 - e^{-(1-\varepsilon_0)v^{*2}t} (1 - (1-\varepsilon_0)(\frac{v^*}{v_0})^2) }}\\
        &\qquad < v <
        v^* \sqrt{\frac{1}{1-e^{-v^{*2}t}(1-(\frac{v^*}{v_0})^2)}} .
    \end{align*}
\end{proof}

\begin{proof}[Proof of Theorem \ref{thm: angle of multi layer}]
    This is almost same with the proof of Theorem \ref{thm: angle of single neuron}.
    Recall that $\varphi(t)$ strictly increases by Theorem \ref{thm: global convergence of multi layer}, thus $\varphi_0 < \varphi(t) < \pi$. From \eqref{eq: angle deep}, we get 
    \begin{align*}
        \frac{\varphi_0}{2\pi} r^{m-1}v^{*m+1}
        &< \csc \varphi \frac{d\varphi}{dt} \\
        &= \frac{\varphi}{2\pi} v^{m-1}v^{*m+1} \\
        &< \frac{1}{2} R^{m-1}v^{*m+1}
    \end{align*}
    
    Integrating both sides, from initial state to time $t$, we get
    \begin{align*}
        \frac{\varphi_0}{2\pi} r^{m-1} v^{*m+1} t
        < \log \frac{\tan \frac\varphi2}{\tan \frac{\varphi_0}{2}} 
        < \frac12 R^{m-1} v^{*m+1}t,
    \end{align*}
    which implies 
    \begin{align*}
        \arctan &\left( \tan \frac{\varphi_0}{2}\; e^{\frac{\varphi_0}{2\pi}r^{m-1}v^{*m+1}t } \right)
        < \frac{\varphi}{2} \\
        &< \arctan \left( \tan \frac{\varphi_0}{2}\; e^{\frac{1}{2}R^{m-1}v^{*m+1}t } \right) .
    \end{align*}
    Using a well-known inequality $\frac\pi2 -\frac1x \le \arctan x \le \frac\pi2 - \frac1x + \frac{1}{3x^3}$, we conclude
    \begin{align*}
        &\frac\pi2 - \cot \frac{\varphi_0}{2} e^{-\frac{\varphi_0}{2\pi}r^{m-1}v^{*m+1}t }
        < \frac{\varphi}{2} \\
        &< \frac\pi2 - \cot \frac{\varphi_0}{2} e^{-\frac{1}{2}R^{m-1}v^{*m+1}t } + \frac13 \cot^3 \frac{\varphi_0}{2} e^{-\frac{3}{2}R^{m-1}v^{*m+1}t }
    \end{align*}
    which completes the proof.
\end{proof}

\subsection{Proof for Section \ref{sec: GD}} \label{app: proof of sec GD}
\begin{proof}[Proof of Theorem \ref{thm: relation between GF and GD}]
    Suppose the gradient flow solution is given by
    \begin{align*}
        w_{GF}(t) = g(e^{-ct}), \qquad w_{GF}(0) = g(1).
    \end{align*}
    Note that its gradient is given by $\frac{dw}{dt} = -ce^{-ct}g'(e^{-ct})$. In another form, it could be written by 
    \begin{align*}
        \frac{dw}{dt}\Big|_{g^{-1}(w)=x} = -cxg'(x), 
    \end{align*}
    since $g$ is one-to-one. Then, we prove the gradient descent solution is given by $w_{GD}(T) = g((1-c\eta)^T) + O(T\eta^2)$, by mathematical induction. Before we start, note that $g(x+O(\eta^2)) = g(x)+O(\eta^2)$ and $g'(x+O(\eta^2)) = g'(x)+O(\eta^2)$, since $g\in C^2([0,1])$ guarantees $g$ and $g'$ are Lipschitz continuous. 
    
    At $T=0$, $w_{GD}(0) = g(1)$. \\
    Suppose $w_{GD}(T) = g((1-c\eta)^T) + O(T\eta^2)$ at step $T$.
    Then,
    \begin{align*}
        & w_{GD}(T+1) \\
        &= w_{GD}(T) + \eta \frac{dw}{dt} \Big|_{w=w_{GD}(T)} \\
        &= w_{GD}(T) + \eta \frac{dw}{dt} \Big|_{w=g((1-c\eta)^T)  +O(T\eta^2)} \\
        &=w_{GD}(T) + \eta \frac{dw}{dt} \Big|_{w=g\left((1-c\eta)^T +O(\eta^2)\right)} \\
        &=w_{GD}(T) + \eta \frac{dw}{dt} \Big|_{g^{-1}(w) = (1-c\eta)^T + O(T\eta^2)} \\
        &= w_{GD}(T) \\
        &\quad -\eta\left[(1-c\eta)^T+O(\eta^2)\right] \; g'(\left[(1-c\eta)^T + O(T\eta^2)\right]) \\
        &= w_{GD}(T) - c\eta (1-c\eta)^T g'((1-c\eta)^T) + O(T\eta^2) \\
        &= g((1-c\eta)^{T}) \\
        &\quad + g'((1-c\eta)^T) \left[ (1-c\eta)^{T+1} - (1-c\eta)^T \right] + O(T\eta^2) \\
        &= g((1-c\eta)^{T+1}) - \frac12 g''(\zeta) [c\eta (1-c\eta)^T]^2 +O(T\eta^2) \\
        &= g((1-c\eta)^{T+1}) + O((T+1)\eta^2).
    \end{align*}
    Note that we use Taylor series expansion $g(x) = g(x_0) + g'(x_0)(x-x_0) + \frac12 g''(\zeta)(x-x_0)^2 $ in the second to last line. This completes the proof of the first part.
    The converse part, approximating gradient flow from gradient descent with form of $g((1-c\eta)^T)$ is accomplished by following above equations in reverse order.
    
    Now, we move to get a universal bound for the error term with further assumptions.
    Suppose $|g^{-1}\ '(y)|<M_1, |g'(x) + xg''(x)|<M_2$, and $|g''(x)| < M_3$ and  $\eta < \frac{1}{cM_1 M_2}$.
    Define $\delta(T) := w_{GD}(T) - g((1-c\eta)^T)$. Then,
    \begin{align*}
        \delta(T+1) &= w_{GD}(T+1) - g((1-c\eta)^{T+1}) \\
        &= w_{GD}(T) + \eta \frac{dw}{dt}\Big|_{w=w_{GD}(T)} \\
        &\qquad -g((1-c\eta)^T(1-c\eta))\\
        &= w_{GD}(T) + \eta \frac{dw}{dt}\Big|_{w=g((1-c\eta)^T)+\delta(T)}\\
        &\qquad- g((1-c\eta)^T -c\eta(1-c\eta)^T) \\
        &= w_{GD}(T) + \eta \frac{dw}{dt}\Big|_{g^{-1}(w)=x} \\
        &\qquad - \Big[g((1-c\eta)^T) - c\eta(1-c\eta)^Tg'((1-c\eta)^T)\\
        &\qquad + \frac12 [c\eta(1-c\eta)^T]^2 g''(\zeta)\Big] \\
        &=\delta(T) - c\eta xg'(x) + c\eta(1-c\eta)^T g'((1-c\eta)^T) \\
        &\qquad -\frac12 [c\eta(1-c\eta)^T]^2 g''(\zeta)
    \end{align*}
    where $x:= g^{-1}(g((1-c\eta)^T)+\delta(T)) = (1-c\eta)^T + \delta(T) g^{-1}\ ' (\xi)$ for some $\xi$ (by MVT). Let $y:=(1-c\eta)^T$. Then, the above equation is continued by
    \begin{align*}
        \delta(T+1) &= \delta(T) - c\eta [xg'(x) -yg'(y)] \\
        &\qquad -\frac12 [c\eta(1-c\eta)^T]^2 g''(\zeta) \\
        &= \delta(T) - c\eta(x-y) [tg'(t)]'|_{t=z} \\
        &\qquad -\frac12 [c\eta(1-c\eta)^T]^2 g''(\zeta) \\
        &= \delta(T) - c\eta \delta(T) g^{-1}\ '(\xi)[g'(z)+zg''(z)] \\
        &\qquad -\frac12 [c\eta(1-c\eta)^T]^2 g''(\zeta) \\
        &= \Big(1-c\eta g^{-1}\ '(\xi)[g'(z)+zg''(z)]\Big)\delta(T) \\
        &\qquad -\frac12 [c\eta(1-c\eta)^T]^2 g''(\zeta).
    \end{align*}
    Therefore, we get
    \begin{align*}
        |\delta(T+1)| &< \Big(1- c\eta M_1 M_2 \Big)|\delta(T)| +\frac12 [c\eta(1-c\eta)^T]^2 M_3 \\
        &< |\delta(T)| + \frac12 c^2\eta^2 M_3 (1-c\eta)^{2T} \\
        &< |\delta(T-1)| +\frac12 c^2\eta^2 M_3 (1-c\eta)^{2T-2} \\
        &\qquad + \frac12 c^2\eta^2 M_3 (1-c\eta)^{2T} \\
        &<\cdots \\
        &<|\delta(0)| + \frac12 c^2\eta^2 M_3 \sum_{k=0}^T (1-c\eta)^{2k} \\
        &< \frac12 c^2\eta^2 M_3 \sum_{k=0}^\infty (1-c\eta)^{2k} \\
        &= \frac12 c^2\eta^2 M_3 \frac{1}{1 - (1-c\eta)^2} \\
        &= \frac{c\eta M_3}{2(2-c\eta)},
    \end{align*}
    which concludes that $\delta(T) = O(\eta)$.
\end{proof}

\begin{proof}[Proof of Theorem \ref{thm: angle of multi layer GD version}]
    Applying Theorem \ref{thm: relation between GF and GD} with $g(x) = \pi - 2\cot \frac{\varphi_0}{2} x$ and $g(x) = \pi - 2\cot \frac{\varphi_0}{2} x + \frac23 \cot^3 \frac{\varphi_0}{2} x^3$ completes the proof. 
\end{proof}

\begin{proof}[Proof of Corollary \ref{cor: required T to estimate angle}]
    For the given $\varepsilon>0$, it is enough to show that the lower bound of Theorem \ref{thm: angle of multi layer GD version} is greater than $\pi-\varepsilon$. Therefore,
    \begin{align*}
        &\quad 2 \cot \frac{\varphi_0}{2} \left(1 -\frac{\varphi_0}{2\pi} r^{m-1} v^{*m+1} \eta \right)^T < \varepsilon \\
        \Leftrightarrow &\quad \left(1 -\frac{\varphi_0}{2\pi} r^{m-1} v^{*m+1} \eta \right)^T < \frac\varepsilon2 \tan \frac{\varphi_0}{2} \\
        \Leftrightarrow &\quad T \log \left(1 -\frac{\varphi_0}{2\pi} r^{m-1} v^{*m+1} \eta \right) < \log \frac\varepsilon2 \tan \frac{\varphi_0}{2} \\
        \Leftrightarrow &\quad T > \frac{\log \frac\varepsilon2 \tan \frac{\varphi_0}{2}}{\log \left(1 -\frac{\varphi_0}{2\pi} r^{m-1} v^{*m+1} \eta \right)}.
    \end{align*}
\end{proof}

\subsection{Proofs for Section \ref{app: GD}} \label{app: proof of app}
\begin{proof}[Proof of Theorem \ref{thm: norm of single neuron GD version}]
    Applying Theorem \ref{thm: relation between GF and GD} with $g(x) = (1-\varepsilon_0)(1-x)\norm{\wb^*}+\norm{\wb_0}x$ and $g(x) = (1-x)\norm{\wb^*} + \norm{\wb_0} x $ immediately proves this theorem. Here, we provide another direct way. 
    
    From \eqref{eq: gradient descent} and \eqref{eq: norm w}, gradient descent method with learning rate $\eta$ at step $T$ is given by
    \begin{align*}
        \norm{\wb(T)} &= \norm{\wb(T-1)} \\
        &\quad +\eta \left(-\frac12 \norm{\wb(T-1)} + \frac12 \norm{\wb^*} (1-\varepsilon) \right) \\
        &= \left(1-\frac{1}{2}\eta\right) \norm{\wb(T-1)} + \frac12 \eta (1-\varepsilon) \norm{\wb^*} .
    \end{align*}
    Using $\varepsilon$ strictly decreases from $\varepsilon_0$ to $0$, we get
    \begin{align*}
        &\frac12 \eta (1-\varepsilon_0) \norm{\wb^*} \\
        &\quad < \norm{\wb(T)} - (1-\frac12 \eta)\norm{\wb(T-1)} \\
        &\quad< \frac12 \eta \norm{\wb^*}.
    \end{align*}
    Telescoping series induces
    \begin{align*}
        &\frac12 \eta \norm{\wb^*} (1-\varepsilon_0) \sum_{k=0}^{T-1} (1-\frac12\eta)^k \\
        &\quad < \norm{\wb(T)} - (1-\frac12 \eta)^{T}\norm{\wb(0)} \\
        &\quad < \frac12 \eta \norm{\wb^*}\sum_{k=0}^{T-1} (1-\frac12\eta)^k.
    \end{align*}
    and thus 
    \begin{align*}
        &\norm{\wb^*} (1-\varepsilon_0) \left[1 - (1-\frac12\eta)^{T} \right] \\
        &\quad< \norm{\wb(T)} - (1-\frac12 \eta)^{T}\norm{\wb(0)} \\
        &\quad < \norm{\wb^*} \left[1 - (1-\frac12\eta)^{T} \right].
    \end{align*}
    Therefore,
    \begin{align*}
        &\norm{\wb^*}(1-\varepsilon_0) - (1-\frac12\eta)^T (\norm{\wb^*}(1-\varepsilon_0)-\norm{\wb(0)}) \\
        &\quad < \norm{\wb(T)} \\
        &\quad < \norm{\wb^*} - (1-\frac12\eta)^T (\norm{\wb^*}-\norm{\wb(0)}).
    \end{align*}
    Finally, if $\norm{\wb(0)}\approx 0$, we get
    \begin{align*}
        &\left[1 - (1-\frac12\eta)^T \right] (1-\varepsilon_0)\norm{\wb^*} \\
        &\quad< \norm{\wb(T)} \\
        &\quad< \left[1 - (1-\frac12\eta)^T \right] \norm{\wb^*},
    \end{align*}
    which completes the proof.
\end{proof}

\begin{proof}[Proof of Theorem \ref{thm: angle of single neuron GD version}]
    Applying Theorem \ref{thm: relation between GF and GD} with $g(x) = \pi - 2\cot \frac{\varphi_0}{2} x$ and $g(x) = \pi - 2\cot \frac{\varphi_0}{2} x + \frac23 \cot^3 \frac{\varphi_0}{2} x^3$ completes the proof. 
\end{proof}

\begin{proof}[Proof of Theorem \ref{thm: norm of two layer GD version}]
    To apply Theorem \ref{thm: relation between GF and GD}, it is enough to check that the function 
    \begin{align*}
        g(x) = \sqrt{\frac{a}{1-(1-b)x}}
    \end{align*}
    is in $C^2([0,1])$ for $a>0$ and $b>0$. A simple computation gives
    \begin{align*}
        g'(x) &= \frac{1-b}{2} \sqrt{\frac{a}{\left( 1-(1-b)x \right)^3}} \\
        g''(x) &= \frac{3(1-b)^2}{4} \sqrt{\frac{a}{\left( 1-(1-b)x \right)^5}},
    \end{align*}
    which are continuous on $[0,1]$.
\end{proof}


\section{Further experimental results}
\subsection{Reconstructed bounds} \label{app: several init}
\begin{figure*}[ht]
	\centering
	\begin{subfigure}[b]{0.45\textwidth}
        \centering
        \includegraphics[width=\textwidth]{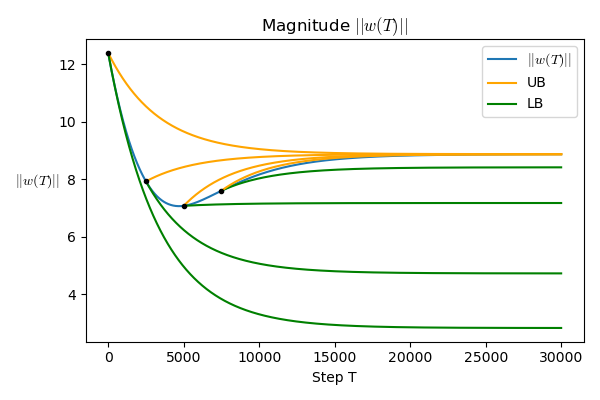}
        \caption{}
    \end{subfigure}
	\hfill
	\begin{subfigure}[b]{0.45\textwidth}
        \centering
        \includegraphics[width=\textwidth]{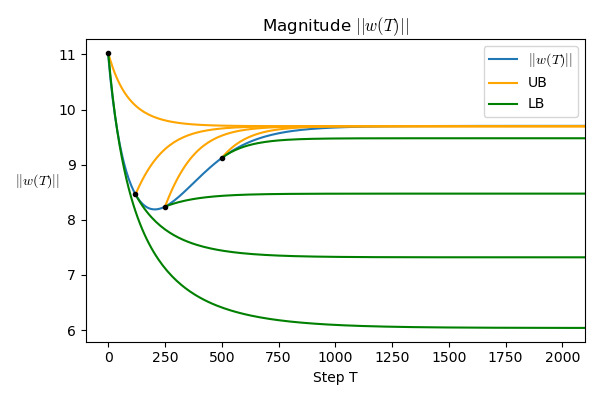}
        \caption{}
    \end{subfigure}
    \caption{
    Upper and lower bounds of $\norm{\wb(T)}$ in one and two-layer single ReLU neurons, on several initialization points.
    (a) According to Theorem \ref{thm: norm of single neuron GD version}, we provide upper and lower bounds (orange, green) for a one-layer single ReLU neuron, initialized at $\wb(0), \wb(2500), \wb(5000)$, and $\wb(7500)$ (black dots).
    (b) Similarly, we provide upper and lower bounds (orange, green)  for a two-layer single ReLU neuron from Theorem \ref{thm: norm of two layer GD version}, initialized at $\wb(0), \wb(120), \wb(250)$, and $\wb(500)$ (black dots).
    }
    \label{fig: several init}
\end{figure*}

Recall the discussion in Section \ref{sec: GF}; Theorem \ref{thm: norm of single neuron} and Theorem \ref{thm: norm of multi layer} provide tighter bounds for smaller $\varepsilon_0$. In this section, we experiment how these bounds are changed during training.
In particular, we compute upper and lower bounds of magnitude at several points on gradient flow for both one and two-layer single ReLU neurons. Detail setting of experiments can be found in Appendix \ref{app: setting}. The result is shown in Figure \ref{fig: several init}. 

For both one and two-layer single ReLU neurons, we can notice that upper and lower bounds are improved as training progresses. Moreover, for each initialization point, the lower bound approximates the front part of gradient flow more accurately, while the upper bound is more accurate for the later part of training. 

We further provide a remark on the convexity of upper and lower bounds. It is easily checked that all bounds proposed in Theorem \ref{thm: norm of single neuron} and Theorem \ref{thm: norm of multi layer} are convex functions. For lower bounds, convexity is changed when $\norm{\wb} = (1-\varepsilon)\norm{\wb^*}$ (one-layer) or $v=(1-\varepsilon)v^*$ (two-layer). Since magnitude usually has two stages; first decreases, and increases to converge, we analyze that the lower bound obtained in the first stage becomes too loose at the second stage. Similarly, convexity of upper bounds is changed when $\norm{\wb} = \norm{\wb^*}$ (one-layer) or $v=v^*$ (two-layer), and we get opposite result in this case. It is inspired to know when $v=(1-\varepsilon)v^*$ occurs, which we leave as an open problem.

\subsection{General deep ReLU networks} \label{app: general deep}

\begin{figure*}[ht]
    \centering
    \begin{subfigure}[b]{0.45\textwidth}
        \centering
        \includegraphics[width=\textwidth]{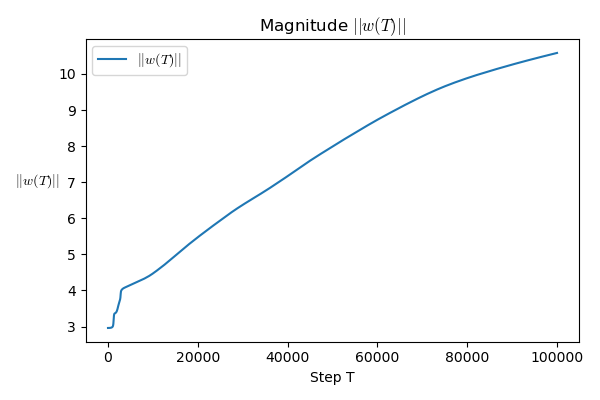}
        \caption{5L-SN}
    \end{subfigure}
    \hfill
    \begin{subfigure}[b]{0.45\textwidth}
        \centering
        \includegraphics[width=\textwidth]{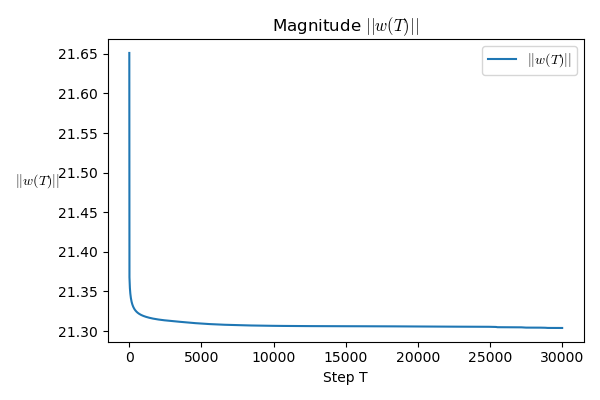}
        \caption{5L-LN}
    \end{subfigure}
    \caption{
    Magnitude dynamics in depth 5 ReLU networks. The left and right graphs represent small and large norm initialization(SN, LN).
    (a) For small norm initialization, magnitude strictly increases until it converges.
    (b) For large norm initialization, magnitude strictly decreases until it converges.
    }
    \label{fig: general deep}
\end{figure*}

In this section, we experimentally check the dynamics of magnitude in general deep ReLU networks (multi-layer with multiple hidden neurons). 
We consider a fully-connected ReLU networks with depth $5$, where all hidden layers have $50$ neurons. Other detail settings can be found in Appendix \ref{app: setting}. The result is shown in Figure \ref{fig: general deep}.
We can easily check that the dynamics of magnitude has monotonicity, for both small and large norm initialization. Note also that small norm initialization indeed slows down convergence speed compare to large norm initialization.

\subsection{Detail setting of experiments} \label{app: setting}
Here we provide detail setting of experiments. 
For all experiments, we construct training dataset $\{\xb_i\}_{i=1}^n \in \Rd^d$ from $\Nc(\zerob, \Ib)$ with $d=100$ and $n=10000$. The label of $\xb_i$ is generated by a target network, as described in Section~\ref{sec: preliminary}. The weight of target network is chosen from $\wb^* \sim \Nc(\zerob, k^*\Ib)$ for each experiment, where $k^*$ is a scaling factor. 
For training network, initialization was chosen from $\wb_0 \sim \Nc(\zerob, k\Ib)$, where $k$ determines initialization scale. 
For chosen $\wb^*$, small norm(SN), middle norm(MN), large norm(LN) initialization refers $k=0.1k^*, k^*$, and $2k^*$, respectively (cf. Figure \ref{fig: angle}, \ref{fig: magnitude}).

If a network is required to be balanced, then initialization of hidden neuron is set to $\norm{\wb_0}$. Since we consider standard gradient descent, we use all data in one update step(iteration/epoch). 
All experiments are implemented by Pytorch.

\paragraph{Experiments in Figure \ref{fig: angle}, \ref{fig: magnitude}}
For one, two, and three-layer single ReLU neurons, we set lr$=0.0003, 0.000008, 0.000002$, and $k^*=1, 0.5, 0.3$, respectively. The total number of iterations (epochs) is set to $100000$. In Theorem \ref{thm: angle of multi layer GD version} and \ref{thm: angle of single neuron GD version}, constants $(r, R)$ are chosen to $(\norm{\wb_0}, \norm{\wb^*})$, $(\norm{\wb^*}/2, \max\{\norm{\wb_0}, \norm{\wb^*}\})$ and $(\norm{\wb^*}/2, \norm{\wb_0})$ for small norm(SN), middle norm(MN), large norm(LN) initialization, respectively.

\paragraph{Experiments in Figure \ref{fig: several init}}
For one, two-layer single ReLU neurons, we set lr=$0.0006$ and $0.0001$, where total iteration numbers(epochs) are set to $30000$ and $2000$, respectively. $k^*$ is set to $1.2$ to observe the dynamics of magnitude. We used four initialization points, which are $\wb(0), \wb(2500), \wb(5000), \wb(7500)$ for one-layer and  $\wb(0), \wb(120), \wb(250), \wb(500)$ for two-layer.

\paragraph{Experiments in Figure \ref{fig: general deep}}
We consider a fully-connected ReLU network, without bias terms, and not balanced. All hidden layers have 50 neurons initialized from $\Nc(\zerob, k\Ib)$, where $k=0.04, 0.44$ for small and large norm initialization, respectively. We set lr=$0.0008, 0.0001$ for (a) and (b) where total iteration numbers(epochs) are set to $100000$ and $30000$.

\end{document}